\documentclass[sigconf]{acmart}

\usepackage{booktabs} 

\setcopyright{rightsretained}

\usepackage{graphicx,epsfig,color}
\usepackage{balance}
\usepackage{comment}
\usepackage{soul}
\usepackage{color}
\usepackage{amsfonts}
\usepackage{url}
\PassOptionsToPackage{hyphens}{url}\usepackage{hyperref}
\usepackage[font={small},labelfont={bf,up}]{caption}
\usepackage[T1]{fontenc}
\usepackage{gensymb}
\usepackage{colortbl}
\usepackage{bm}
\usepackage{multirow}
\usepackage{subcaption}
\usepackage{amsmath}
\usepackage{bm}
\usepackage[normalem]{ulem} 
\newcommand{\sysname}{DeepASL}
\newcommand{\leap}{Leap Motion}

\newcommand{\modelone}{HB-RNN}

\newcommand{\squishlist}{
   \begin{list}{$\bullet$}
    { \setlength{\itemsep}{0pt}      \setlength{\parsep}{3pt}
      \setlength{\topsep}{3pt}       \setlength{\partopsep}{0pt}
      \setlength{\leftmargin}{1.5em} \setlength{\labelwidth}{1em}
      \setlength{\labelsep}{0.5em} } }
\newcommand{\squishend}{
    \end{list}  }

\begin{document}

\copyrightyear{2017} 
\acmYear{2017} 
\setcopyright{acmcopyright}
\acmConference{SenSys '17}{November 6--8, 2017}{Delft, Netherlands}\acmPrice{15.00}\acmDOI{10.1145/3131672.3131693}
\acmISBN{978-1-4503-5459-2/17/11}

\title[{\sysname}]{\textit{{\sysname}}: Enabling Ubiquitous and Non-Intrusive Word and Sentence-Level Sign Language Translation}


\author{Biyi Fang, Jillian Co, Mi Zhang}
\affiliation{%
  \institution{Michigan State University}
}

\renewcommand{\shortauthors}{B. Fang et al.}

\begin{abstract}

There is an undeniable communication barrier between deaf people and people with normal hearing ability.
%
Although innovations in sign language translation technology aim to tear down this communication barrier, the majority of existing sign language translation systems are either intrusive or constrained by resolution or ambient lighting conditions.
Moreover, these existing systems can only perform single-sign ASL translation rather than sentence-level translation, making them much less useful in daily-life communication scenarios.  
In this work, we fill this critical gap by presenting \textit{{\sysname}}, a transformative deep learning-based sign language translation technology that enables ubiquitous and non-intrusive American Sign Language (ASL) translation at both word and sentence levels.
{\sysname} uses infrared light as its sensing mechanism to non-intrusively capture the ASL signs.
It incorporates a novel hierarchical bidirectional deep recurrent neural network (HB-RNN) and a probabilistic framework based on Connectionist Temporal Classification (CTC) for word-level and sentence-level ASL translation respectively. 
%
%
To evaluate its performance, we have collected $7,306$ samples from $11$ participants, covering $56$ commonly used ASL words and $100$ ASL sentences.
{\sysname} achieves an average $94.5\%$ word-level translation accuracy and an average $8.2\%$ word error rate on translating unseen ASL sentences.
%
%
Given its promising performance, we believe {\sysname} represents a significant step towards breaking the communication barrier between deaf people and hearing majority, and thus has the significant potential to fundamentally change deaf people's lives.

%


\end{abstract}

%
%



\maketitle

\newpage


\section{Introduction}
\label{sec.intro}

\begin{figure*}[tb]
\vspace{-2mm}
\centering
\includegraphics[scale=0.55]{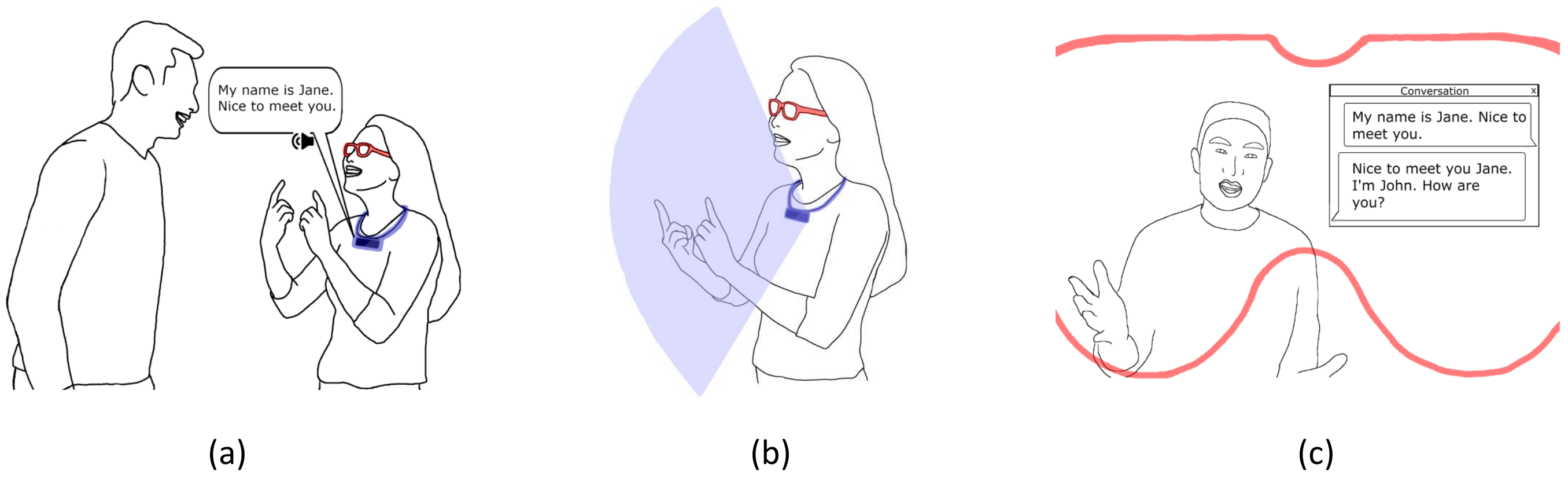}
\vspace{-1.0mm}
\caption{Illustration of an envisioned scenario of real-time two-way communication enabled by {\sysname}: (a) {\sysname} translates the signs performed by the deaf person into spoken English and broadcasts the translated ASL sentence via a speaker; (b) {\sysname} captures the signs in a non-intrusive manner; (c) {\sysname} leverages the speech recognition technology to translate spoken English into texts, and projects the texts through a pair of augmented reality (AR) glasses.}
\vspace{-3.5mm}
\label{fig:vision}
\end{figure*} 

\noindent
%
In the United States, there are over 28 million people considered deaf or hearing disabled~\cite{deafstats1}.
American Sign Language, or ASL in short, is the primary language used by deaf people to communicate with others~\cite{deafstats2}.
Unfortunately, very few people with normal hearing understand sign language.
Although there are a few methods for aiding a deaf person to communicate with people who do not understand sign language, such as seeking help from a sign language interpreter, writing on paper, or typing on a mobile phone, each of these methods has its own key limitations in terms of cost, availability, or convenience.
%
As a result, there is an undeniable communication barrier between deaf people and hearing majority.


%
At the heart of tearing down this communication barrier is the sign language translation technology.
Sign language is a language like other languages but based on signs rather than spoken words.
A sign language translation system uses sensors to capture signs and computational methods to map the captured signs to English.
%
Over the past few decades, although many efforts have been made, sign language translation technology is still far from being practically useful.
Specifically, existing sign language translation systems use motion sensors, Electromyography (EMG) sensors, RGB cameras, Kinect sensors, or their combinations~\cite{starner1998real, brashear2003using, wu2015real, li2012sign, chai2013visualcomm} to capture signs. 
Unfortunately, these systems are either intrusive where sensors have to be attached to fingers and palms of users, lack of resolutions to capture the key characteristics of signs, or significantly constrained by ambient lighting conditions or backgrounds in real-world settings.
%
More importantly, existing sign language translation systems can only translate a single sign at a time, thus requiring users to pause between adjacent signs.
These limitations significantly slow down face-to-face conversations, making those sign language translation systems much less useful in daily-life communication scenarios.


%
In this paper, we present \textit{{\sysname}}, a transformative deep learning-based sign language translation technology that enables non-intrusive ASL translation at both word and sentence levels.
%
%
{\sysname} can be embedded inside a wearable device, a mobile phone, a tablet, a laptop, a desktop computer, or a cloud server to enable ubiquitous sign language translation.
As such, {\sysname} acts as an \textit{always-available virtual sign language interpreter}, which allows deaf people to use their primary language to communicate with the hearing majority in a natural and convenient manner.
As an example, Figure~\ref{fig:vision} illustrates an envisioned scenario where {\sysname} is in the form of a wearable device, enabling a deaf person and a hearing individual who does not understand ASL to use their own primary languages to communicate with each other face to face.
%
Specifically, from one side, {\sysname} translates signs performed by the deaf person into spoken English; from the other side, {\sysname} leverages the speech recognition technology to translate English spoken from the hearing individual into text, and projects the text through a pair of augmented reality (AR) glasses for the deaf person to read.


%
{\sysname} uses {\leap}~\cite{leapmotion} -- an infrared light-based sensing device that can extract the skeleton joints information of fingers, palms and forearms -- to non-intrusively capture the ASL signs performed by a deaf person. 
By leveraging the extracted skeleton joints information, {\sysname} achieves word and sentence-level ASL translation via three innovations. 
First, {\sysname} leverages domain knowledge of ASL to extract the key characteristics of ASL signs buried in the raw skeleton joints data.
Second, {\sysname} employs a novel hierarchical bidirectional deep recurrent neural network (HB-RNN) to effectively model the spatial structure and temporal dynamics of the extracted ASL characteristics for word-level ASL translation.
%
%
Third, {\sysname} adopts a probabilistic framework based on Connectionist Temporal Classification (CTC)~\cite{graves2006connectionist} for sentence-level ASL translation.
%
This eliminates the restriction of pre-segmenting the whole sentence into individual words, and thus enables translating the whole sentence end-to-end directly  without requiring users to pause between adjacent signs. 
%
%
Moreover, it enables {\sysname} to translate ASL sentences that are not included in the training dataset, and hence eliminates the burden of collecting all possible ASL sentences.


\vspace{1.5mm}
\noindent
\textbf{Summary of Experimental Results:}
We have conducted a rich set of experiments to evaluate the performance of {\sysname} in three aspects: 
1) ASL translation performance at both word level and sentence level; 
%
2) robustness of ASL translation under various real-world settings; 
and 3) system performance in terms of runtime, memory usage and energy consumption. 
Specifically, to evaluate the ASL translation performance, we have collected $7,306$ samples from $11$ participants, covering $56$ commonly used ASL words and $100$ ASL sentences.
To evaluate the robustness, we have collected $1,178$ samples under different ambient lighting conditions, body postures when performing ASL, and scenarios with in-the-scene interference and multi-device interference. 
%
%
To evaluate the system performance, we have implemented {\sysname} on three platforms with different computing power: 1) a desktop equipped with an Intel i7-4790 CPU and a Nvidia GTX 1080 GPU (desktop CPU and GPU), 2) a Nvidia Jetson TX1 mobile development board equipped with an ARM Cortex-A57 CPU and a Nvidia Tegra X1 GPU (mobile CPU and GPU), and 3) a Microsoft Surface Pro 4 tablet equipped with an Intel i5-6300 CPU (tablet CPU).
Our results show that:
\squishlist{
\item{
At the word level, {\sysname} achieves an average $94.5\%$ translation accuracy. 
At the sentence level, {\sysname} achieves an average $8.2\%$ word error rate on translating unseen ASL sentences and an average $16.1\%$ word error rate on translating ASL sentences performed by unseen users. 
} 
\item{
{\sysname} achieves more than $91.8\%$ word-level ASL translation accuracy in various ambient lighting conditions, body postures, and interference sources, demonstrating its great robustness in real-world daily communication scenarios.
}

\item{{\sysname} achieves $282$ ms in runtime performance in the worst-case scenario across three platforms for both word-level and sentence translation. 
It also demonstrates the capability of supporting enough number of inferences for daily usage on both mobile and tablet platforms.  

}
}\squishend

\vspace{1.5mm}
\noindent
\textbf{Summary of Contributions:}
%
The development of sign language translation technology dates back to the beginning of 90s~\cite{von2008recent}.
However, due to the limitations in both sensing technology and computational methods, limited progress has been made over the decades.
%
The innovative solution provided by {\sysname} effectively addresses those limitations, and hence represents a significant contribution to the advancement of sign language translation technology.
%
Moreover, the development of {\sysname} enables a wide range of applications. 
As another contribution of this work, we have designed and developed two prototype applications on top of {\sysname} to demonstrate its practical value. 
According to World Health Organization (WHO), there are an estimated 360 million people worldwide having disabling hearing loss~\cite{who}.
While the focus of this paper is on American sign language translation, since our approach is generic at modeling signs expressed by hands, it can be leveraged for developing sign language translation technologies for potentially any of the three hundred sign languages in use around the world~\cite{signlanguagelist}.
Given its promising performance, we believe {\sysname} represents a significant step towards breaking the communication barrier between deaf people and hearing majority, and thus has the significant potential to fundamentally change deaf people's lives.

\newpage

\section{Background, State-of-the-Art, and Design Choice}
\label{sec.background}
\vspace{-2mm}
\noindent


\subsection{Characteristics of ASL}
\begin{figure}[h]
\centering
\includegraphics[scale=0.171]{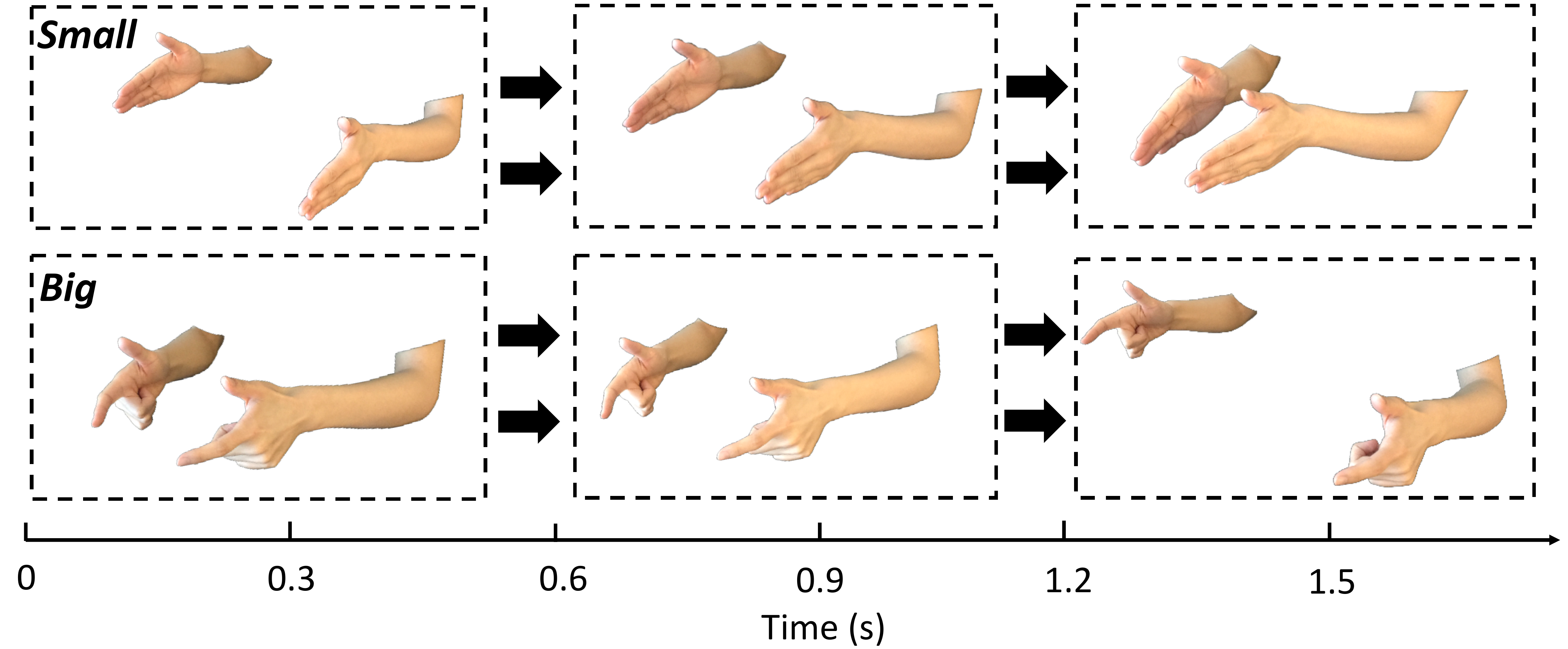}
\vspace{-1mm}
\caption{Illustration on how hand shape, hand movement, and relative location of two hands altogether encodes the meaning of two ASL signs: ``small'' and ``big''.}
\vspace{-2mm}
\label{fig:asl-hand}
\end{figure}

ASL is a complete and complex language that mainly employs signs made by moving the hands~\cite{hoza2007s}.
%
Each individual sign is characterized by three key sources of information: 1) hand shape, 2) hand movement, and 3) relative location of two hands \cite{hoza2007s, liddell2003grammar}. 
It is the combination of these three key characteristics that encodes the meaning of each sign.
As an example, Figure~\ref{fig:asl-hand} illustrates how these three characteristics altogether encode the meaning of two ASL signs: ``small'' and ``big''. 
Specifically, to sign ``small'', one starts with holding both hands in front of her with fingers closed (i.e., hand shape), and then moves two hands towards each other (i.e., hand movement and relative location).
In comparison, to sign ``big'', one starts with extending the thumb and index fingers to form a slightly bent 'L' shape (i.e., hand shape), and then moves two hands away from each other (i.e., hand movement and relative location). 

It is worthwhile to note that, for illustration purpose, we have selected two of the most distinctive ASL signs to explain the characteristics of ASL. 
%
In fact, there are many ASL signs that involve very subtle differences in the three key characteristics mentioned above. 
Moreover, in real-world scenarios, ASL signs can be expressed under various conditions such as bright vs. poor lighting conditions, walking vs. standing; and indoor vs. outdoor environments.
It is the subtle differences and the real-world factors altogether that makes the task of ASL translation challenging.


\subsection{State-of-the-Art ASL Translation Systems}

Based on the sensing modality the system uses, existing ASL translation systems can be generally grouped into four categories: 1) wearable sensor-based, 2) Radio Frequency (RF)-based, 3) RGB camera-based, and 4) Kinect-based systems.
However, each of them has fundamental limitations that prevent it from being practically useful for translating ASL in daily life scenarios.
Specifically,
wearable sensor-based systems \cite{kim2008bi, kosmidou2009sign, li2010automatic, li2012sign, wu2015real, tubaiz2015glove, praveen2014sign, abhishek2016glove, kanwal2014assistive} use motion sensors (accelerometers, gyroscopes), EMG sensors, or bend sensors to capture the movements of hands, muscle activities, or bending of fingers to infer the performed signs.
However, wearable sensor-based systems require attaching sensors to a user's fingers, palms, and forearms.
This requirement makes them very intrusive and impractical for daily usage.
RF-based systems \cite{melgarejo2014leveraging} use wireless signals as a sensing mechanism to capture hand movements. 
Although this contactless sensing mechanism minimizes the intrusiveness to users, wireless signals have very limited resolutions to ``see'' the hands.
%
RGB camera-based systems \cite{starner1998real, zafrulla2010novel, brashear2003using}, on the other hand, are capable of capturing rich information about hand shape and hand movement without instrumenting users. 
However, they fail to reliably capture those information in poor lighting conditions or generally uncontrolled backgrounds in real-world scenarios.
Moreover, the videos/images captured may be considered invasive to the privacy of the user and surrounding bystanders.
Finally, although Kinect-based systems overcome the lighting and privacy issues of the RGB camera-based systems by only capturing the skeleton information of the user body and limbs \cite{chai2013sign, chai2013visualcomm}, they do not have enough resolution to capture the hand shape information, which plays a critical role on decoding the sign language.

\subsection{Design Choice}

\begin{figure}[h]
\centering
\includegraphics[scale=0.166]{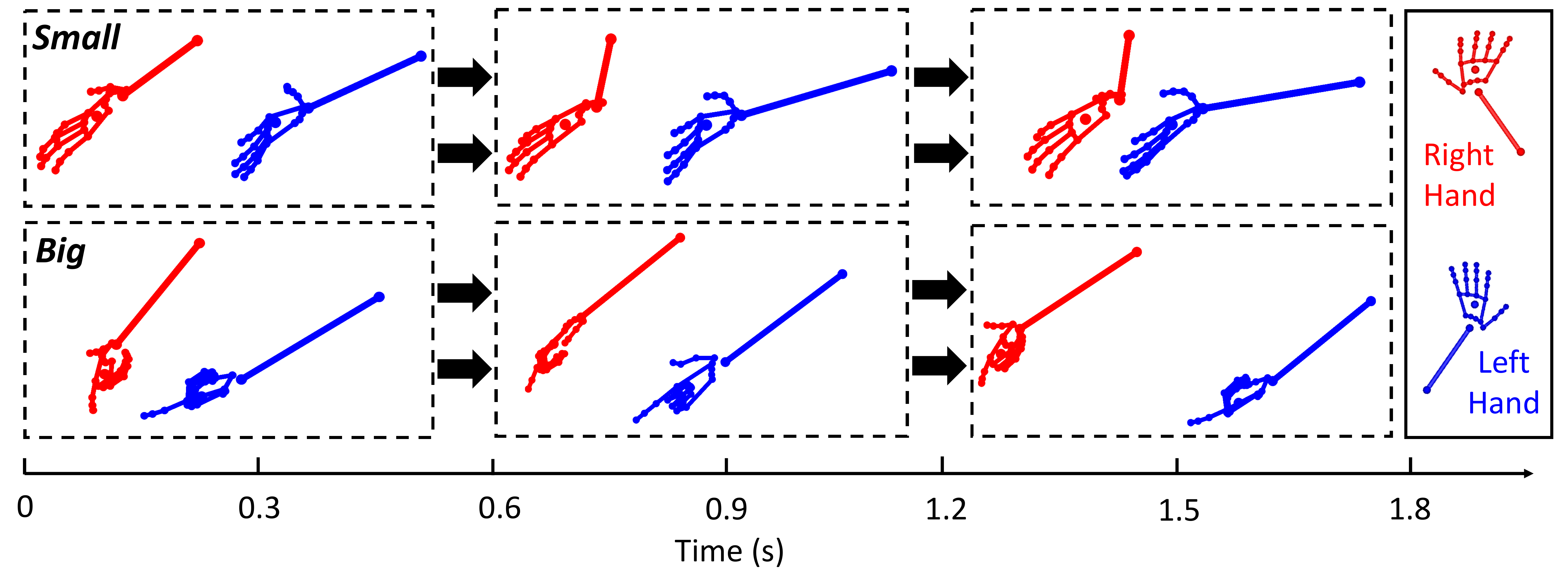}
\vspace{-6.5mm}
\caption{The skeleton joints of two ASL signs: ``small'' and ``big''.}
\vspace{-1.5mm}
\label{fig:asl-leapmotion}
\end{figure}

In the design of {\sysname}, we use {\leap} as our sensing modality to capture ASL signs~\cite{leapmotion}. 
{\leap} overcomes the fundamental limitations of existing technologies and is able to precisely capture the three key characteristics of ASL signs under real-world scenarios in a non-intrusive manner. 
%
Specifically, {\leap} uses infrared light as its sensing mechanism.
This not only enables it to capture the signs in a contactless manner but also makes it ``see'' the signs in poor lighting conditions.
Moreover, {\leap} is able to extract skeleton joints of the fingers, palms and forearms from the raw infrared images.
%
This preserves the privacy of the user and bystanders, and more importantly, provides enough resolution to precisely capture hand shape as well as hand movements and locations.
As an example, Figure~\ref{fig:asl-leapmotion} illustrates how the ASL signs of two words ``small'' and ``big'' are precisely captured by the temporal sequence of skeleton joints of the fingers, palms and forearms. 

\vspace{-1mm}


To sum up, Table~\ref{tab.comp} compares {\leap} with other sensing modalities used in existing sign language translation systems. 
As listed, {\leap} has shown its superiority over other sensing modalities on capturing the three key characteristics of ASL signs in a non-intrusive manner without the constraint of ambient lighting condition.
We leverage this superiority in the design of {\sysname}.

\begin{table}[h]	
\begin{center}
\scalebox{0.71}{
\begin{tabular}{|c|c|c|c|c|c|}
\hline
\textbf{Sensing}    	& \textbf{Hand}         	& \textbf{Hand}                  & \textbf{Hand}        	& \textbf{Intrusive}    & \textbf{Lighting}     \\ 
\textbf{Modality}    	& \textbf{Shape}         	& \textbf{Movement}          & \textbf{Location}    & \textbf{}			& \textbf{Condition}  \\ \hline
Motion + EMG + Bend      	& Captured 		& Captured 			& No                          & Yes                        & Any                        \\ \hline
RF                   		& No      				& Captured 			& No 			& No 			& Any                        \\ \hline
RGB Camera           	& Captured                   	& Captured                        & Captured               & Yes                        & Constrained  		\\ \hline
Kinect                 	& No      				& Captured 			& Captured 		& No 			& Any                       \\ \hline
Leap Motion 		& Captured          		& Captured            		& Captured               & No                          & Any                         \\ \hline
\end{tabular}
 }
 \vspace{0.5mm}
\caption{Comparison of sensing modalities for ASL translation.}
\vspace{-7.5 mm}
\label{tab.comp}
\end{center}
\end{table}

\newpage

\section{Challenges and Our Solutions}
\label{sec.challenges}
\noindent
Although {\leap} has shown its superiority over other sensing modalities on capturing key characteristics of ASL signs, there is a \textit{significant gap} between the raw skeleton joints data and the translated ASL.
%
In this section, we describe the challenges on transforming the raw skeleton joints data into translated ASL at both word and sentence levels. 
We also explain how {\sysname} effectively addresses those challenges.



\vspace{1.0mm}
\noindent
\textbf{ASL Characteristics Extraction:}
{\leap} is \textit{not} designed for ASL translation.
Although {\leap} captures the skeleton joints of the fingers, palms and forearms, the key information that characterizes ASL signs (i.e., hand shape, hand movement, and relative location of two hands) is still buried in the raw skeleton joints data.
%
To address this challenge, we leverage domain knowledge of ASL to extract spatio-temporal trajectories of ASL characteristics from the sequence of skeleton joints during signing, and develop models upon the extracted ASL characteristics for ASL translation.


\vspace{1.0mm}
\noindent
\textbf{ASL Characteristics Organization:} 
The extracted ASL characteristics are isolated and unorganized, and thus can not be directly used for ASL translation. 
This problem is exacerbated when the number of ASL signs to be translated scales up.
To address this challenge, we propose a hierarchical model based on deep recurrent neural network (RNN) that effectively integrates the isolated low-level ASL characteristics into an organized high-level representation that can be used for ASL translation.  



\begin{figure}[h]
\centering
\includegraphics[scale=0.318]{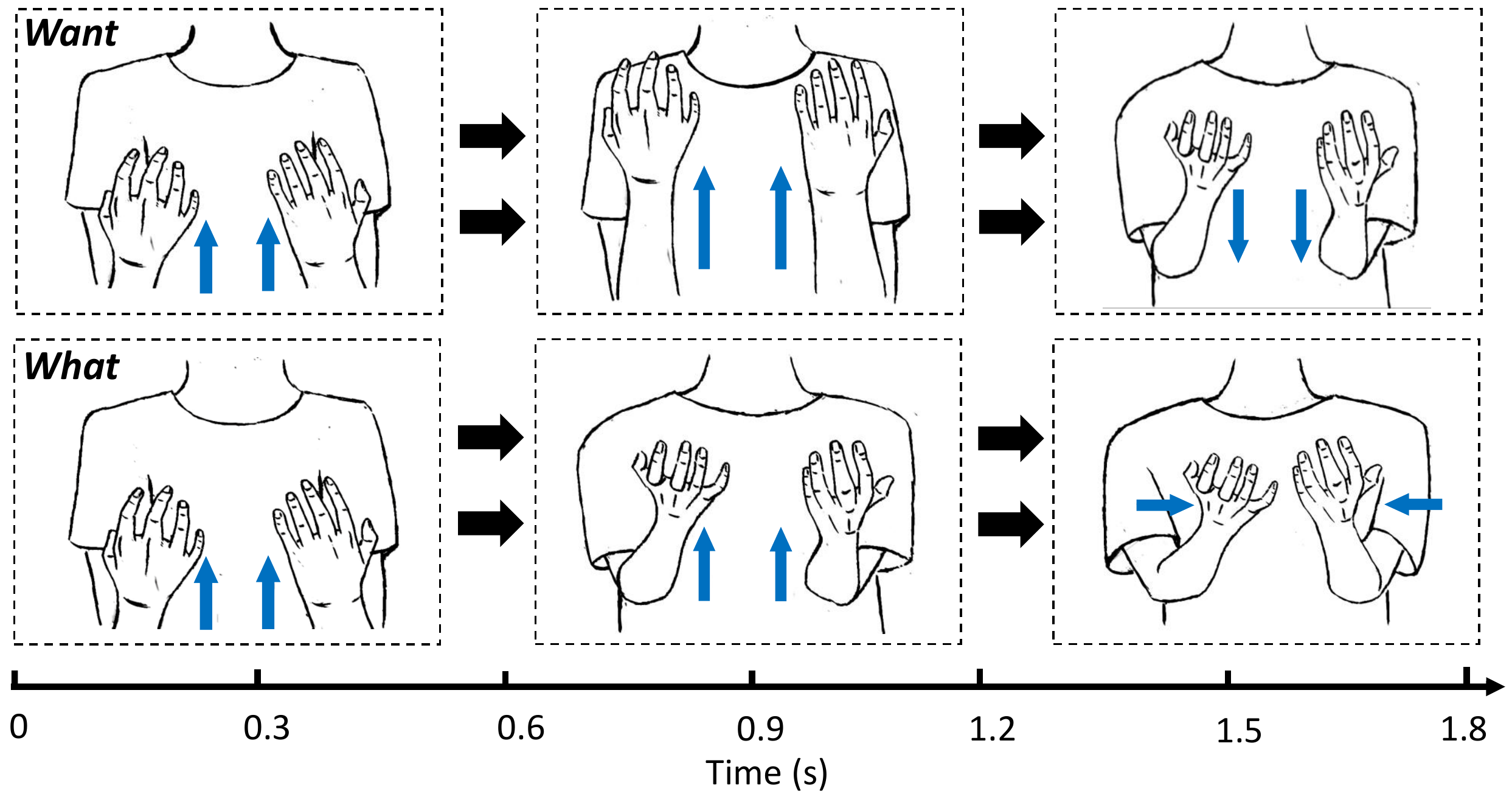}
\vspace{-3mm}
\caption{Similarity between two ASL signs: ``want'' and ``what''.}
\vspace{-2mm}
\label{fig:similarity}
\end{figure}

\vspace{1.0mm}
\noindent
\textbf{Similarity between Different Signs:} 
Although each ASL sign is uniquely characterized by its ASL characteristics trajectories, many ASL signs share very similar characteristics at the beginning of their trajectories (see Figure~\ref{fig:similarity} as an example). 
%
This similarity confuses traditional RNN which is based on a unidirectional architecture.
This is because the unidirectional architecture can only use the past information at each time point in the trajectory to infer the sign being performed.
To address this challenge, we propose a bidirectional RNN model which performs inference at each point of the trajectory based on both past and future trajectory information. 
%
With the global view of the entire trajectory, our bidirectional RNN model is able to achieve better ASL translation performance. 

\vspace{1.0mm}
\noindent
\textbf{ASL Sentence Translation:}
%
To translate ASL sentences, existing sign language translation technologies adopt a framework which requires pre-segmenting individual words within the sentence. 
However, this framework restricts sign language translation technologies to translate one single sign at a time and thus requires users to pause between adjacent signs when signing one sentence.
%
To address this challenge, we propose to adopt a framework based on Connectionist Temporal Classification (CTC) that computes the probability of the whole sentence directly, and therefore, removes the requirement of pre-segmentation.


%
To the best of our knowledge, {\sysname} is the first ASL translation framework that addresses these challenges and achieves accurate ASL translation performance at word and sentence levels.


\section{System Overview}
\label{sec.overview}
\begin{figure}[t]
\centering
\includegraphics[scale=0.37]{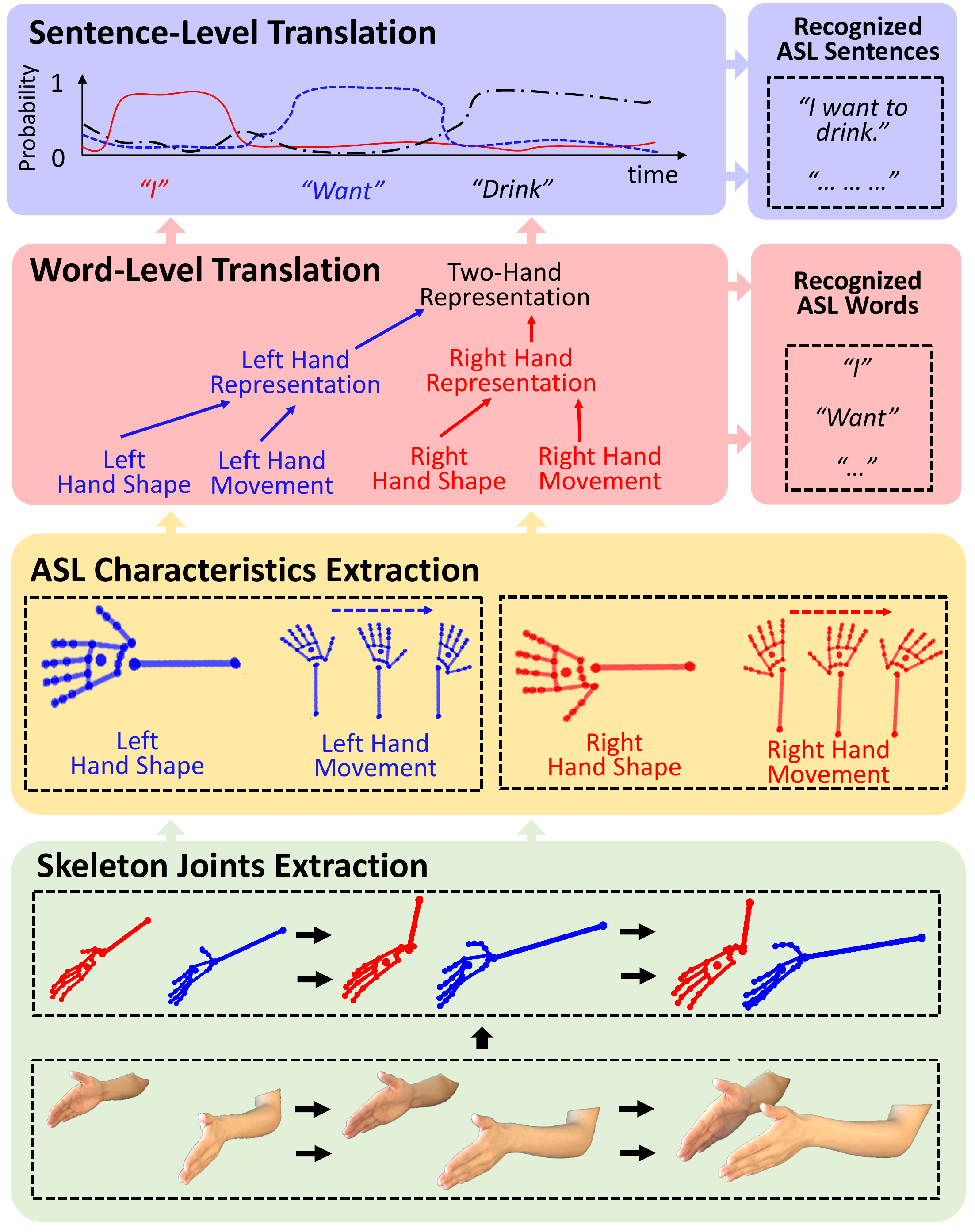}
\caption{The system architecture of {\sysname}.}
\vspace{-4mm}
\label{dia.sys}
\end{figure}

\noindent
%
%
Figure~\ref{dia.sys} provides an overview of the multi-layer system architecture of {\sysname}.
%
Specifically, at the first layer, a temporal sequence of 3D coordinates of the skeleton joints of fingers, palms and forearms is captured by the {\leap} sensor during signing.
%
At the second layer, the key characteristics of ASL signs including hand shape, hand movement and relative location of two hands are extracted from each frame of the sequence, resulting in a number of spatio-temporal trajectories of ASL characteristics.
%
At the third layer, {\sysname} employs a hierarchical bidirectional deep recurrent neural network (HB-RNN) that models the spatial structure and temporal dynamics of the spatio-temporal trajectories of ASL characteristics for word-level ASL translation.
%
Finally, at the top layer, {\sysname} adopts a CTC-based framework that leverages the captured probabilistic dependencies between words in one complete sentence and translates the whole sentence end-to-end without requiring users to pause between adjacent signs.
%
In the next section, we describe the design of {\sysname} in details.



\newpage

\section{System Details}
\label{sec.algorithm}

\subsection{ASL Characteristics Extraction}
\begin{figure}[h]
\centering
\includegraphics[scale=0.47]{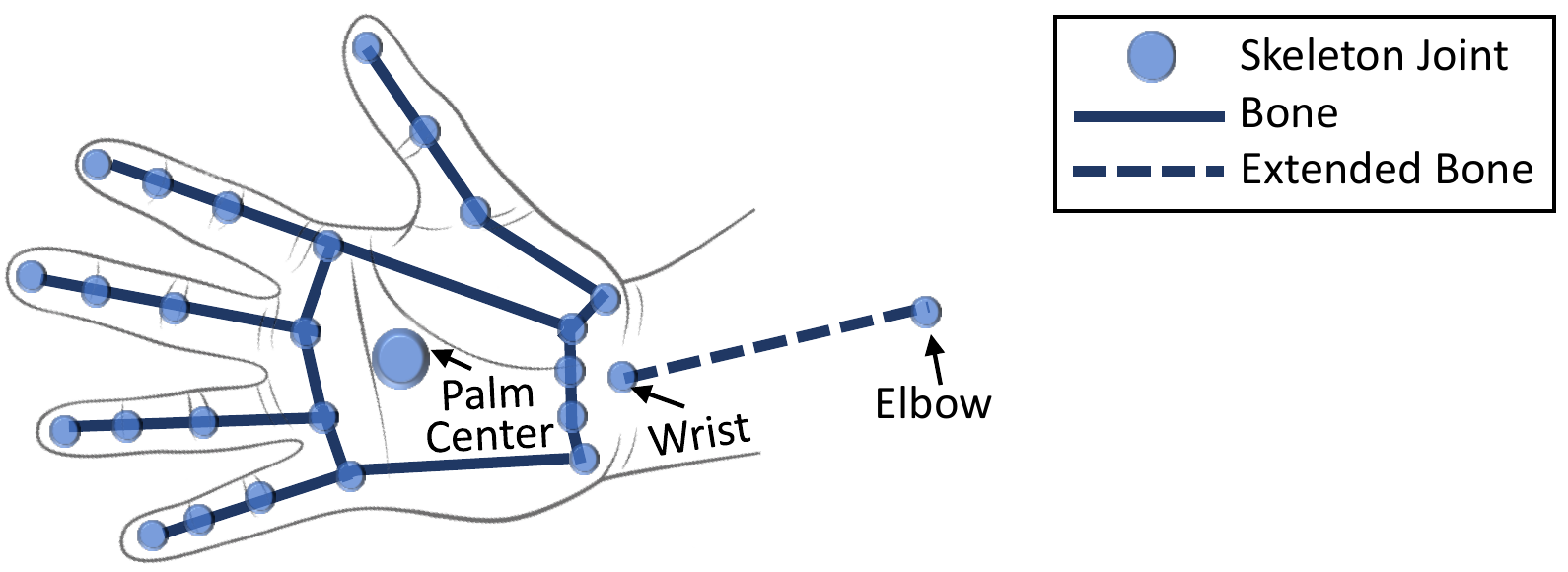}
\vspace{-2mm}
\caption{The skeleton joints tracked by the Leap Motion sensor.}
\vspace{-2mm}
\label{fig:skeleton}
\end{figure}

The skeleton joints data provided by the {\leap} sensor is noisy in its raw form. 
%
As our first step, we apply a simple Savitzky-Golay filter~\cite{savitzky1964smoothing} to improve the signal to noise ratio of the raw skeleton joints data.
We select the Savitzky-Golay filter because of its effectiveness in smoothing skeleton joints data~\cite{Zhu2016,du2015hierarchical}.
%
Specifically, let $J_{i,j,t} = (x_{i,j,t}, y_{i,j,t}, z_{i,j,t}),~i = \{left, right\},~j = \{1, ..., N\},~t = \{1, ..., T\}$ denote the $t$-th frame of the temporal sequence of the 3D coordinates of the skeleton joints of fingers, palms and forearms of a single ASL sign, where $x$, $y$, $z$ denote the 3D coordinates of the skeleton joints, $i$ is the hand index, $j$ is the skeleton joint index (see Figure~\ref{fig:skeleton} for the skeleton joints tracked by the Leap Motion sensor), $t$ is the frame index, $N$ denotes the total number of skeleton joints in one hand, and $T$ denotes the total number of frames included in the temporal sequence.
%
The Savitzky-Golay filter is designed as
\begin{equation} 
\widetilde{J}_{i,j,t} = (-3J_{i,j,t-2} + 12J_{i,j,t-1} + 17J_{i,j,t} + 12J_{i,j,t+1} - 3J_{i,j,t+2})/35
\end{equation}
where $\widetilde{J}_{i,j,t}$ denotes the smoothed 3D coordinates of the skeleton joints in the $t$-th frame.

\vspace{-3mm}


%
Based on the smoothed temporal sequence of the skeleton joints data, we extract the key characteristics of ASL signs including hand shape, hand movement and relative location of two hands from each frame of the sequence.
Specifically, since hand shape is independent of the absolute spatial location of the hand and is characterized by the relative distances among skeleton joints of palm and fingers, we extract hand shape information of both left and right hands by zero-centering the palm center of the right hand and then normalizing the 3D coordinates of the skeleton joints to it as
\begin{equation} 
S_{i,j,t} = \widetilde{J}_{i,j,t} - \widetilde{J}_{i,j = right\_palm\_center,t.}
\vspace{1.5mm}
\end{equation}
%
By doing this, the information of the relative location of the left hand to the right hand is also encoded in $S_{i=left,j,t}$. 
Lastly, we extract hand movement information of both left and right hands as the spatial displacement of each skeleton joint between two consecutive frames defined as
\begin{equation}
\vspace{-0.5mm}
  M_{i,j,t}=\begin{cases}
    (0, 0, 0), & \text{if $t = 1$} \\
    \widetilde{J}_{i,j,t} - \widetilde{J}_{i,j,t-1}, & \text{if $t = 2, ..., T$}.
      \end{cases}
  \vspace{1.5mm}
\end{equation}

Taken together, the ASL characteristics extracted from each frame of the temporal sequence of 3D coordinates of the skeleton joints result in four spatio-temporal ASL characteristics trajectories that capture information related to: 1) right hand shape, 2) right hand movement, 3) left hand shape (it also encodes the information of the relative location of the left hand to the right hand), and 4) left hand movement, respectively.
We denote them as $\bm{S}_{right}$, $\bm{M}_{right}$, $\bm{S}_{left}$, and $\bm{M}_{left}$ accordingly.

\subsection{Word-Level ASL Translation}

In this section, we first provide the background knowledge of bidirectional recurrent neural network (B-RNN) and Long Short-Term Memory (LSTM) to make the paper self-contained. 
We then describe our proposed hierarchical bidirectional deep recurrent neural network (HB-RNN) which is designed upon B-RNN and LSTM for single-sign word-level ASL translation. 
Finally, we describe the architectures of four comparative models that we use to validate the design choice of our proposed model.



\vspace{2mm}
\subsubsection{A Primer on Bidirectional RNN and LSTM}
~\newline
RNN is a powerful model for sequential data modeling~\cite{graves2012supervised}.
%
It has been widely used and has shown great success in many important tasks such as speech recognition~\cite{graves2013speech}, natural language processing~\cite{socher2011parsing}, language translation~\cite{sutskever2014sequence}, and video recognition~\cite{donahue2015long}.
Specifically, given an input temporal sequence $\textbf{x} = (x_{1}, x_{2}, ..., x_{T})$, where in our case $x_{t}$ is the $t$-th frame of the spatio-temporal ASL characteristics trajectories, the hidden states of a recurrent layer $\textbf{h} = (h_{1}, h_{2}, ..., h_{T})$ and the output $\textbf{y} = (y_{1}, y_{2}, ..., y_{T})$ of a RNN can be obtained as:
\begin{equation} 
\label{equ.rnn1}
h_{t} = \theta_{h} (W_{xh} x_{t} + W_{hh} h_{t-1} + b_{h})
\vspace{0.5mm}
\end{equation}
\begin{equation}
\label{equ.rnn2}
y_{t} =\theta_{y} (W_{ho} h_{t} + b_{o})
\vspace{0.5mm}
\end{equation}
where $W_{xh}$, $W_{hh}$, and $W_{ho}$ are connection weight matrices, $b_h$ and $b_o$ are bias values, and $\theta_h$ and  $\theta_y$ are activation functions.

\begin{figure}[b]
\centering
\includegraphics[scale=0.348]{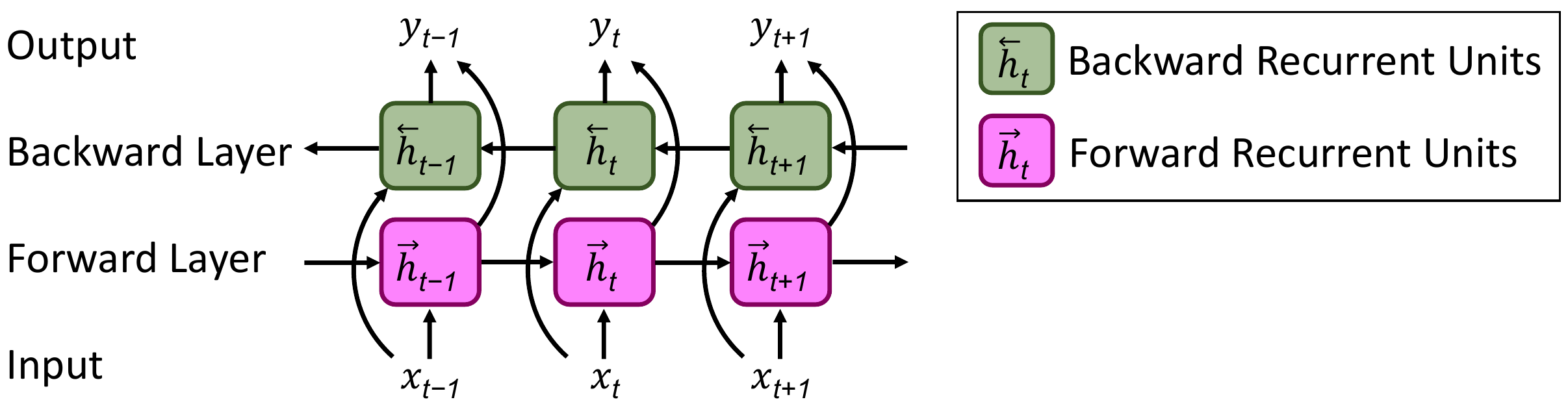}
\vspace{-2mm}
\caption{The network architecture of bidirectional RNN (B-RNN).}
\vspace{-2mm}
\label{fig:b-rnn}
\end{figure} 

The RNN model described above only contains a single recurrent hidden layer and is unidirectional.
One limitation of the unidirectional RNN is that it could only look backward and thus can only access the past information at each time point in the temporal sequence for inference.
%
In the context of sign language translation, this limitation could cause translation errors when different signs share very similar characteristics at the beginning of the signs. 
To address this limitation, we propose to use bidirectional RNN (B-RNN)~\cite{schuster1997bidirectional} as the building block in our design. 
Figure~\ref{fig:b-rnn} illustrates the network architecture of a B-RNN.
As shown, B-RNN has two separate recurrent hidden layers, with one pointing backward (i.e., backward layer) and the other pointing forward (i.e., forward layer).
As such, a B-RNN can look both backward and forward, and can thus utilize both the past and future information at each time point in the temporal sequence to infer the sign being performed.

The recurrent structure of RNN enables it to learn complex temporal dynamics in the temporal sequence. 
However, it can be difficult to train a RNN to learn long-term dynamics due to the vanishing and exploding gradients problem~\cite{hochreiter2001gradient}.
%
To solve this problem, Long Short-Term Memory (LSTM)~\cite{hochreiter1997long} was invented which enables the network to learn when to forget previous hidden states and when to update hidden states given new input.
This mechanism makes LSTM very efficient at capturing the long-term dynamics. 
%
Given this advantage, we use B-RNN with LSTM architecture in our design to capture the complex temporal dynamics during signing.

\begin{figure}[t]
\centering
\includegraphics[scale=0.23]{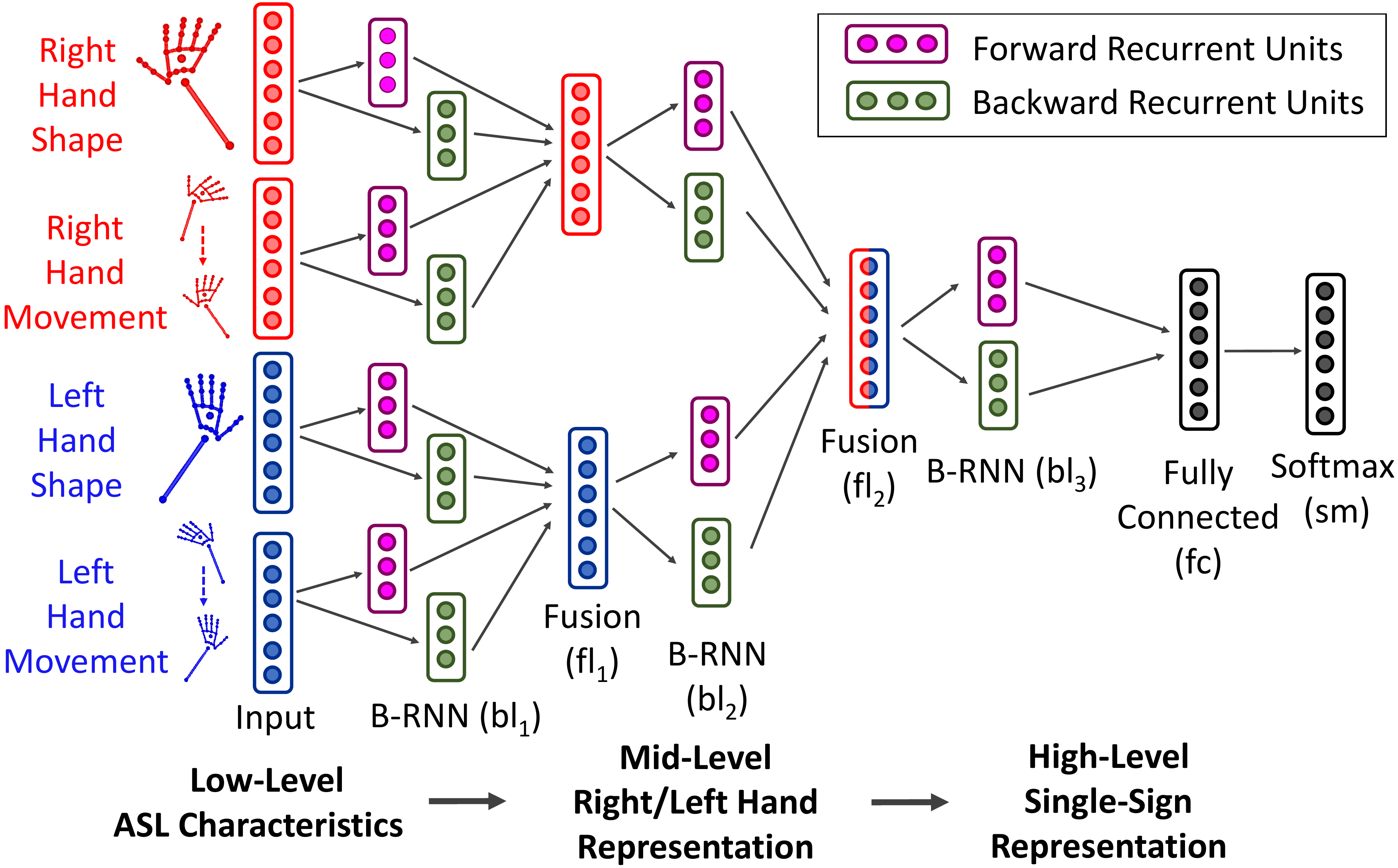}
\caption{The architecture of the hierarchical bidirectional deep recurrent neural network (HB-RNN) for word-level ASL translation. For ASL signs that are performed using only one hand, only the corresponding half of the {\modelone} model is activated.}
\vspace{-2mm}
\label{dia.HB-RNN}
\end{figure}

\vspace{2mm}
\subsubsection{Hierarchical Bidirectional RNN for Single-Sign Modeling}
~\newline
Although four spatio-temporal ASL characteristics trajectories have been extracted from the raw skeleton joints data, they are isolated and at low level, and thus can not be directly used for word-level ASL translation.
Therefore, we propose a hierarchical model based on bidirectional deep recurrent neural network with the LSTM architecture (HB-RNN) to integrate the isolated low-level ASL characteristics into an organized high-level representation that can be used for word-level ASL translation.

Figure~\ref{dia.HB-RNN} illustrates the architecture of the proposed HB-RNN model. 
At a high level, our HB-RNN model takes the four spatio-temporal ASL characteristics trajectories as its input, extracts the spatial structure and the temporal dynamics within the trajectories, and combines them in a hierarchical manner to generate an integrated high-level representation of a single ASL sign for word-level ASL translation.
As shown, our {\modelone} model consists of seven layers including three B-RNN layers ($bl_{1,2,3}$), two fusion layers ($fl_{1,2}$), one fully connected layer ($fc$), and one softmax layer ($sm$). 
Each of these layers has different structure and thus plays different role in the whole model.
Specifically, in the $bl_{1}$ layer, the four spatio-temporal ASL characteristics trajectories that capture information related to the right hand shape ($\bm{S}_{right}$), right hand movement ($\bm{M}_{right}$), left hand shape ($\bm{S}_{left}$), and left hand movement ($\bm{M}_{left}$) are fed into four separate B-RNNs.
%
These B-RNNs capture the spatial structure among skeleton joints and transform the low-level ASL characteristics into new representations of right hand shape, right hand movement, left hand shape, and left hand movement in both forward layer $\overrightarrow{h}$ and backward layer $\overleftarrow{h}$.
%
In the fusion layer $fl_{1}$, we concatenate the newly generated representations of right (left) hand shape and right (left) hand movement together as 
%
$R_{bl_{1}}^{i,t} = \lbrace\overrightarrow{h}_{bl_{1}}^{t}(\bm{S}^{t}_{i}),\overleftarrow{h}_{bl_{1}}^{t}(\bm{S}^{t}_{i}),\overrightarrow{h}_{bl_{1}}^{t}(\bm{M}^{t}_{i}),\overleftarrow{h}_{bl_{1}}^{t}(\bm{M}^{t}_{i}) \rbrace,~~i=\{right,left\},$
and feed the two concatenations into two B-RNNs in the $bl_{2}$ layer separately to obtain an integrated representation of the right (left) hand.
%
Similarly, the two newly generated right and left hand representations are further concatenated together in the fusion layer $fl_{2}$ denoted as $R_{bl_{2}}^{t}$.
This concatenation is then fed into the B-RNN in the $bl_{3}$ layer to obtain a high-level representation in both forward layer and backward layer (denoted as $\overrightarrow{h}_{bl_{3}}^{t}(R^{t}_{bl_{2}})$ and $\overleftarrow{h}_{bl_{3}}^{t}(R^{t}_{bl_{2}})$) that integrates all the ASL characteristics of a single sign.  
%
Finally, we connect $\overrightarrow{h}_{bl_{3}}^{t}(R^{t}_{bl_{2}})$ and $\overleftarrow{h}_{bl_{3}}^{t}(R^{t}_{bl_{2}})$ to the fully connected layer $fc$.
The output of $fc$ is summed up across all the frames in the temporal sequence and then normalized by the softmax function in the softmax layer $sm$ to calculate the predicted word class probability given a sequence $J$:
 \begin{equation} 
\label{equ.sum}
O =  \sum\limits_{t=1}^T O_{fc}^{t}
\vspace{-1.2mm}
\end{equation}
\vspace{-0.4mm}
\begin{equation} 
\label{equ.sm}
p(C_{k} | J) =  \frac{e^{O_{k}}}{\sum\nolimits_{n=1}^{C}e^{O_{n}}}, k = 1, ..., C
\vspace{1.5mm}
\end{equation}
where $C$ denotes the total number of ASL words in the dictionary.
%
By accumulating results and normalizing across all the frames, our model is able to make inference based on the information of the entire sequence.
More importantly, it allows our model to handle ASL signs that have different sequence lengths as well as sequence length variation caused by signing speed.

\vspace{2mm}
\subsubsection{Comparative Models}
~\newline
To validate the design choice of our proposed HB-RNN model, we construct four comparative models as follows:

\squishlist{
\item{
\textbf{HB-RNN-M:} a hierarchical bidirectional RNN model with hand movement information only. We compare this model with HB-RNN to prove the importance of the hand shape information.
} 
\item{
\textbf{HB-RNN-S:} a hierarchical bidirectional RNN model with hand shape information only. We compare this model with HB-RNN to prove the importance of the hand movement information.
}

\item{\textbf{SB-RNN:} a simple bidirectional RNN model without hierarchical structure. We compare this model with HB-RNN to prove the importance of the hierarchical structure.
}
\item{\textbf{H-RNN:} a hierarchical unidirectional RNN model with forward recurrent layer only. We compare this model with HB-RNN to prove the significance of the bidirectional connection.
}
}\squishend

The parameters of the proposed HB-RNN model as well as the four comparative models are listed in Table~\ref{tab.parameters}. 

\begin{table}[h]
\centering
\scalebox{0.82}{
\begin{tabular}{|c|c|c|c|c|}
\hline
\textbf{Category}                                                                    & \textbf{Model} & \textbf{RNN Layer 1}                  & \textbf{RNN Layer 2}                 & \textbf{RNN Layer 3}                 \\ \hline
\multirow{5}{*}{\begin{tabular}[c]{@{}c@{}}One-Hand\\ ASL\\ Words\end{tabular}}  & HB-RNN-M              & 2 $\times$ 1 $\times$ 128  & -                        & -                        \\ \cline{2-5} 
                                                                            & HB-RNN-S              & 2 $\times$ 1 $\times$ 128  & -                        & -                        \\ \cline{2-5} 
                                                                            & SB-RNN                 & 2 $\times$ 1 $\times$ 128 & -                        & -                        \\ \cline{2-5} 
                                                                            & H-RNN                 & 1 $\times$ 2 $\times$ 64  & 1 $\times$ 1 $\times$ 128 & -                        \\ \cline{2-5} 
                                                                            & {\modelone}          & 2 $\times$ 2 $\times$ 32  & 2 $\times$ 1 $\times$ 64 & -                        \\ \hline
\multirow{5}{*}{\begin{tabular}[c]{@{}c@{}}Two-Hand\\ ASL\\ Words\end{tabular}} & HB-RNN-M              & 2 $\times$ 2 $\times$ 64  & 2 $\times$ 1 $\times$ 128 & -                        \\ \cline{2-5} 
                                                                            & HB-RNN-S              & 2 $\times$ 2 $\times$ 64  & 2 $\times$ 1 $\times$ 128 & -                        \\ \cline{2-5} 
                                                                            & SB-RNN                 & 2 $\times$ 1 $\times$ 256 & -                        & -                        \\ \cline{2-5} 
                                                                            & H-RNN                 & 1 $\times$ 4 $\times$ 64  & 1 $\times$ 2 $\times$ 64 & 1 $\times$ 1 $\times$ 128 \\ \cline{2-5} 
                                                                            &{\modelone}          & 2 $\times$ 4 $\times$ 32  & 2 $\times$ 2 $\times$ 32 & 2 $\times$ 1 $\times$ 64 \\ \hline
\end{tabular}
}
\vspace{2mm}
 \caption{The parameters of our proposed HB-RNN model and the four comparative models. The parameters follow the format of 1 (unidirectional) or 2 (bidirectional) $\times$  \#RNNs $\times$ \#hidden units. 
 }
 \vspace{-7mm}
 \label{tab.parameters}
\end{table}


\subsection{Sentence-Level ASL Translation}
\label{sec.algorithm2}
In daily-life communication, a deaf person does not sign a single word but a complete sentence at a time.
Although the {\modelone} model described in the previous section is capable of transforming the low-level ASL characteristics into a high-level representation for word-level translation, when translating a complete ASL sentence, {\modelone} still requires pre-segmenting the whole sentence into individual words and then connecting every translated word into a sentence in the post-processing. 
This is not only complicated but also requires users to pause between adjacent signs when signing one sentence, which is not practical in daily-life communication. 

To address this problem, we propose a probabilistic approach based on Connectionist Temporal Classification (CTC)~\cite{graves2006connectionist} for sentence-level ASL translation.
%
CTC is the key technique that drives the modern automatic speech recognition systems such as Apple Siri and Amazon Alexa~\cite{mcgraw2016personalized}.
It eliminates the necessity of word pre-segmentation and post-processing, allowing end-to-end translation of a whole sentence.
%
Inspired by its success on sentence-level speech recognition, we propose a CTC-based approach that can be easily built on top of the {\modelone} model described in the previous section for sentence-level ASL translation.
%
Specifically, to realize sentence-level ASL translation based on CTC, we make the following modifications on {\modelone}:
%
\squishlist{
\item{
Let $V$ denote the ASL word vocabulary.
%
We add a blank symbol $\{blank\}$ into the ASL word vocabulary: $V'= V\cup\{blank\}$.
%
Essentially, this blank symbol enables us to model the transition from one word to another within a single sentence. 
}
\item{
We increase the capacity of the RNN layers (i.e., $bl_1$, $bl_2$ and $bl_3$) in HB-RNN to 2 $\times$ 4 $\times$ 32, 2 $\times$ 2 $\times$ 64, and 2 $\times$ 1 $\times$ 128, respectively (see Table \ref{tab.parameters} for the format definition). 
This is because ASL sentences are more complex than ASL words and thus require more parameters for modeling.
}
\item{
%
Since an ASL sentence consists of multiple signs, we replace the softmax layer in HB-RNN which computes the probability of a single sign with a new softmax layer which computes the probabilities of a sequence of multiple signs.
%
}
\item{
Based on the modified softmax layer, the probabilities of all the possible sentences formed by the word included in $V$ can be computed.
Given those probabilities, we compute the probability of a target label sequence by marginalizing over all the sequences that are defined as equivalent to this sequence. 
For example, the label sequence $'SL'$ is defined as equivalent to the label sequences $'SSL'$, $'SLL'$, $'S\underline{\hspace{2mm}} L'$ or $'SL\underline{\hspace{2mm}}'$, where $'\underline{\hspace{2mm}}'$ denotes the blank symbol $\{blank\}$. 
This process not only eliminates the need for word pre-segmentation and post-processing but also addresses variable-length sequences.
}
%
\item{
Finally, we delete adjacent duplicate labels and remove all the blank symbols in the inferred label sequence to derive the translated sentence. 
}
}\squishend

With all the above modifications, the end-to-end sentence-level ASL translation is achieved.
\newpage

\section{Evaluation}
\label{sec.evaluation}


\subsection{Experimental Setup}
\vspace{1mm}
\subsubsection{Dataset Design}
~\newline
To evaluate the translation performance of {\sysname} at both word and sentence levels as well as its robustness under real-world settings, we have designed and collected three datasets: 
1) ASL Words Dataset; 2) ASL Sentences Dataset; and 3) In-the-Field Dataset. 
 
\vspace{1mm}
\noindent
\textbf{ASL Words Dataset}: 
%
%
Since it is impossible to collect all the words in the ASL vocabulary, we target ASL words that are representative of each category of the ASL vocabulary.
In particular, we have selected 56 ASL words from five word categories: pronoun, noun, verb, adjective and adverb.
Table~\ref{tab.aslwords} lists the selected 56 ASL words. 
%
These words are among the top 200 most commonly used words in ASL vocabulary. 
Among these 56 words, 29 are performed by two hands and the rest 27 are performed by one hand (right hand). 

\vspace{-1mm}
\begin{table}[h]
\centering
\scalebox{0.92}{
\begin{tabular}{|c|p{6.5cm}|}
\hline
\textbf{Category} & \hspace{25mm} \textbf{Words}                      \\ \hline
pronoun           & who, I, you, \underline{what}, we, my, your, other                       \\ \hline
noun              & \underline{time}, food, drink, mother, \underline{clothes}, \underline{box}, \underline{car}, \underline{bicycle}, \underline{book}, \underline{shoes}, \underline{year}, boy, \underline{church}, \underline{family}  \\ \hline
verb              & \underline{want}, \underline{dontwant}, like, \underline{help}, \underline{finish}, need, thankyou, \underline{meet}, \underline{live}, \underline{can}, come \\ \hline
adjective         & \underline{big}, \underline{small}, hot, \underline{cold}, blue, red, \underline{gray}, black, green, white, old, \underline{with}, \underline{without}, \underline{nice}, bad, \underline{sad}, \underline{many}, sorry, few         \\ \hline
adverb            & where, \underline{more}, please, \underline{but}                     \\ \hline
\end{tabular}
}
\vspace{2mm}
\caption{The ASL Words Dataset (two-hand words are underlined).}
\vspace{-7mm}
\label{tab.aslwords}
\end{table} 
 
 
\vspace{1mm}
\noindent
\textbf{ASL Sentences Dataset}:
%
%
We have followed the dataset design methodology used in Google's LipNet (i.e., sentence-level lipreading) \cite{assael2016lipnet} to design our ASL Sentences Dataset.
Specifically, we design the ASL sentences by following a simple sentence template: $subject^{(4)} + predicate^{(4)} +  attributive^{(4)} + object^{(4)}$, where the superscript denotes the number of word choices for each of the four word categories, which are designed to be $\lbrace I, you, mother, who \rbrace$, $\lbrace dontwant, like, want, need \rbrace$, $\lbrace big, small, cold, more \rbrace$ and $\lbrace time, \\ food, drink, clothes \rbrace$, respectively.
%
%
Based on this sentence template, a total of $256$ possible sentences can be generated. 
%
Out of these 256 sentences, we hand picked $100$ meaningful sentences that people would use in daily communication.
Example meaningful sentences are "I need more food" and "Who want cold drink".
%
 

\vspace{1mm}
\noindent
\textbf{In-the-Field Dataset}: 
%
In daily life, deaf people may need to use ASL to communicate with others under various real-world settings.
We consider three common real-world factors that can potentially affect the ASL translation performance: 1) lighting conditions, 2) body postures; and 3) interference sources. 
For lighting conditions, we collected data from both indoor poor lighting scenario and outdoor bright sunlight scenario. 
%
For body postures, we collected data when signs are performed while the signer stands or walks.
For interference sources, we considered two interference sources: people and device.
In terms of people interference, data was collected while another person stands in front of {\leap} with both of her hands appearing in the viewing angle of the {\leap} sensor. 
This setup simulates the communication scenario between a deaf person and a normal hearing person. 
In terms of device interference, data was collected while another person is wearing {\leap} standing near the user. 
%
This setup simulates the communication scenario where there are more than one person using {\sysname}.

\vspace{1mm}
\subsubsection{Participants}
~\newline
%
Our study is approved by IRB.
Due to the IRB constraint, we could only recruit people with normal hearing ability to participate in the study.
%
We recruited 11 participants and hosted an 3-hour tutorial session to teach them how to perform the target ASL signs using online ASL tutorial videos.
%
The 11 participants (four female) are between 20 to 33 ($\mu$ = 24.2) years old, weighted between 49 kg to 86 kg ($\mu$ = 74 kg) and are between 155 cm to 185 cm tall ($\mu$ = 173 cm).

\vspace{1mm}
\subsubsection{Summary of Datasets}
~\newline
Table~\ref{tab.sum} summarizes the amount of data collected in the three datasets.
Specifically, for ASL Words Dataset, we collected 56 ASL words with 10 ($\pm$3) samples of each word from each of the 11 participants. 
In total, $3,068$ and $3,372$ samples of one-hand and two-hand ASL words were collected, respectively.
For ASL Sentences Dataset, we randomly collected 80 ($\pm$3) out of the $100$ meaningful sentences from each of the 11 participants.
In total, $866$ sentences were collected.
For In-the-Field Dataset, for each of the six scenarios, we randomly selected 25 out of the 56 ASL words and collected 3 ($\pm$1) samples of each word from three out of the 11 participants.
To the best of our knowledge, our datasets are the largest and the most comprehensive datasets in the sign language translation literature~\cite{kosmidou2009sign,dominio2014combining,uebersax2011real,zafrulla2010novel,melgarejo2014leveraging,brashear2003using}. 

%
%



\vspace{-2mm}
\begin{table}[h]
\centering
\scalebox{0.83}{
\begin{tabular}{|c|c|c|c|c|c|}
\hline
\textbf{Category}    & \multicolumn{2}{c|}{\textbf{ASL Words}} & \multirow{2}{*}{\begin{tabular}[c]{@{}c@{}}\textbf{ASL}\\\textbf{Sentences} \end{tabular}} & \multirow{2}{*}{\begin{tabular}[c]{@{}c@{}} \textbf{In-the-} \\\textbf{Field}\end{tabular}} & \multirow{2}{*}{\textbf{Total}} \\ \cline{1-3}
Subcategory &\textbf{One-hand}           &\textbf{Two-hand}          &                                                                                    &                                                                                                       &                        \\ \hline
Duration (s)  &       7541.7           &    8616.3       &             5094.3
 &           2498.4                                                                                            &                       23750.7 \\ \hline
Frames      &     821846        &      949310     &                                                                                   507001 &                 259431         &       2537588      \\ \hline
Samples     &     
3068      &   3372   &          866                &                                                                                                      1178 &        8484         \\ \hline
\end{tabular}
}
\vspace{0mm}
\caption{Summary of datasets}
\vspace{-8.5mm}
\label{tab.sum}
\end{table}


\subsubsection{Evaluation Metrics and Protocol}
\vspace{1mm}
\noindent
\textbf{\\Evaluation Metrics:} 
We use different metrics to evaluate the translation performance of {\sysname} at the word and sentence levels.
Specifically, at the word level, we use word translation accuracy, confusion matrix, and Top-$K$ accuracy as evaluation metrics. 
%
%
At the sentence level, we use word error rate ($WER$) as the evaluation metric, which is defined as the minimum number of word insertions, substitutions, and deletions required to transform the prediction into the ground truth, divided by the number of words in the ground truth. 
$WER$ is also the standard metric for evaluating sentence-level translation performance of speech recognition systems.
%
%


\vspace{1mm}
\noindent
\textbf{Evaluation Protocol:} 
At both word and sentence levels, we use leave-one-subject-out cross-validation as the protocol to examine the generalization capability of {\sysname} across different subjects. 
%
In addition, to evaluate the performance of {\sysname} in translating unseen sentences, we randomly divide ASL sentences into ten folds, making sure that each fold contains unique sentences to the rest nine folds. 
We then use ten-fold cross-validation as the protocol. 
%


\begin{figure}[t]
\centering
\hspace{-6mm}
\includegraphics[scale=0.53]{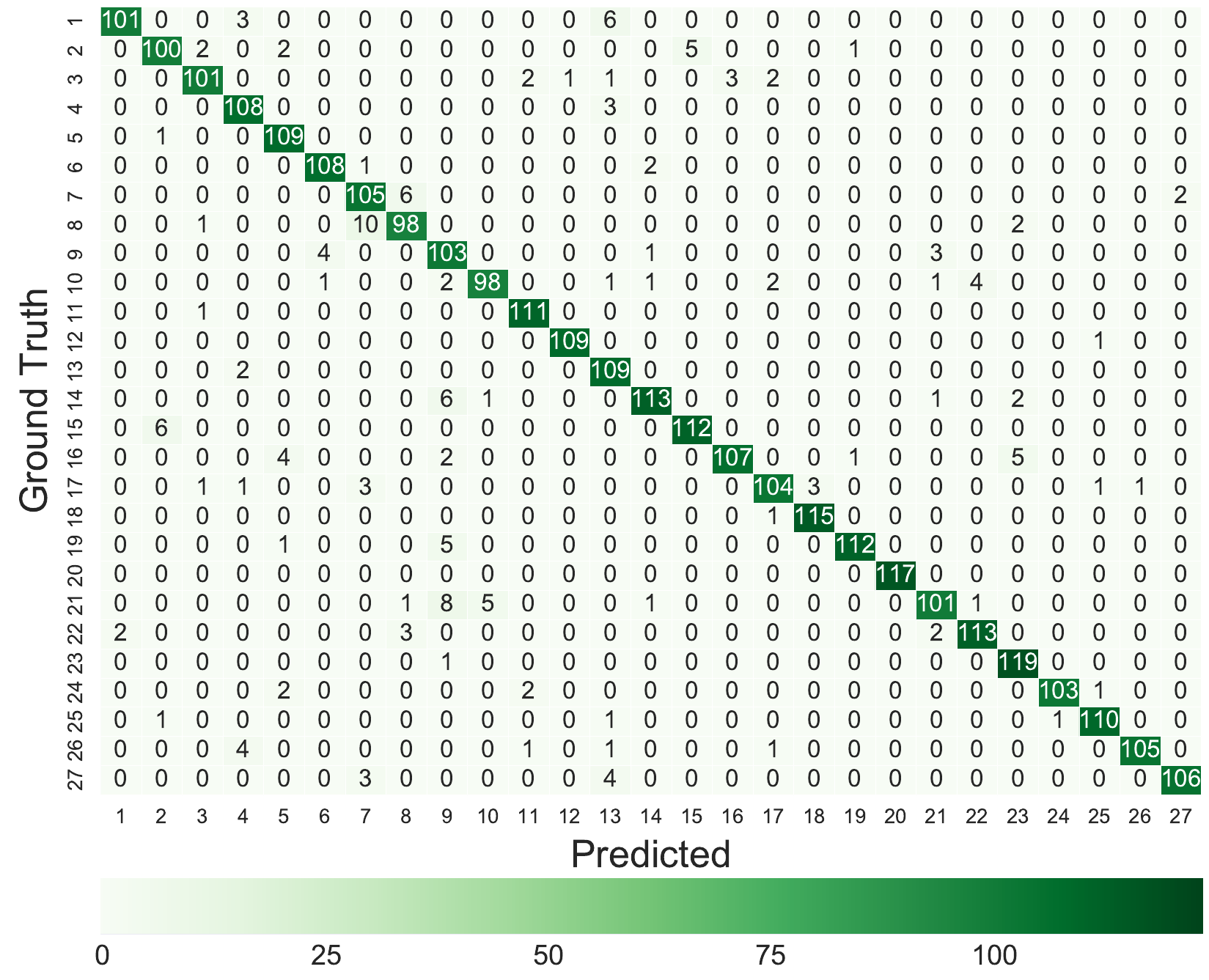}
\vspace{-2mm}
\caption{Confusion matrix of 27 one-hand ASL words.}
\vspace{-1mm}
\label{dia.confusion27}
\end{figure} 

\begin{figure}[h]
\centering
\hspace{-5mm}
\includegraphics[scale=0.52]{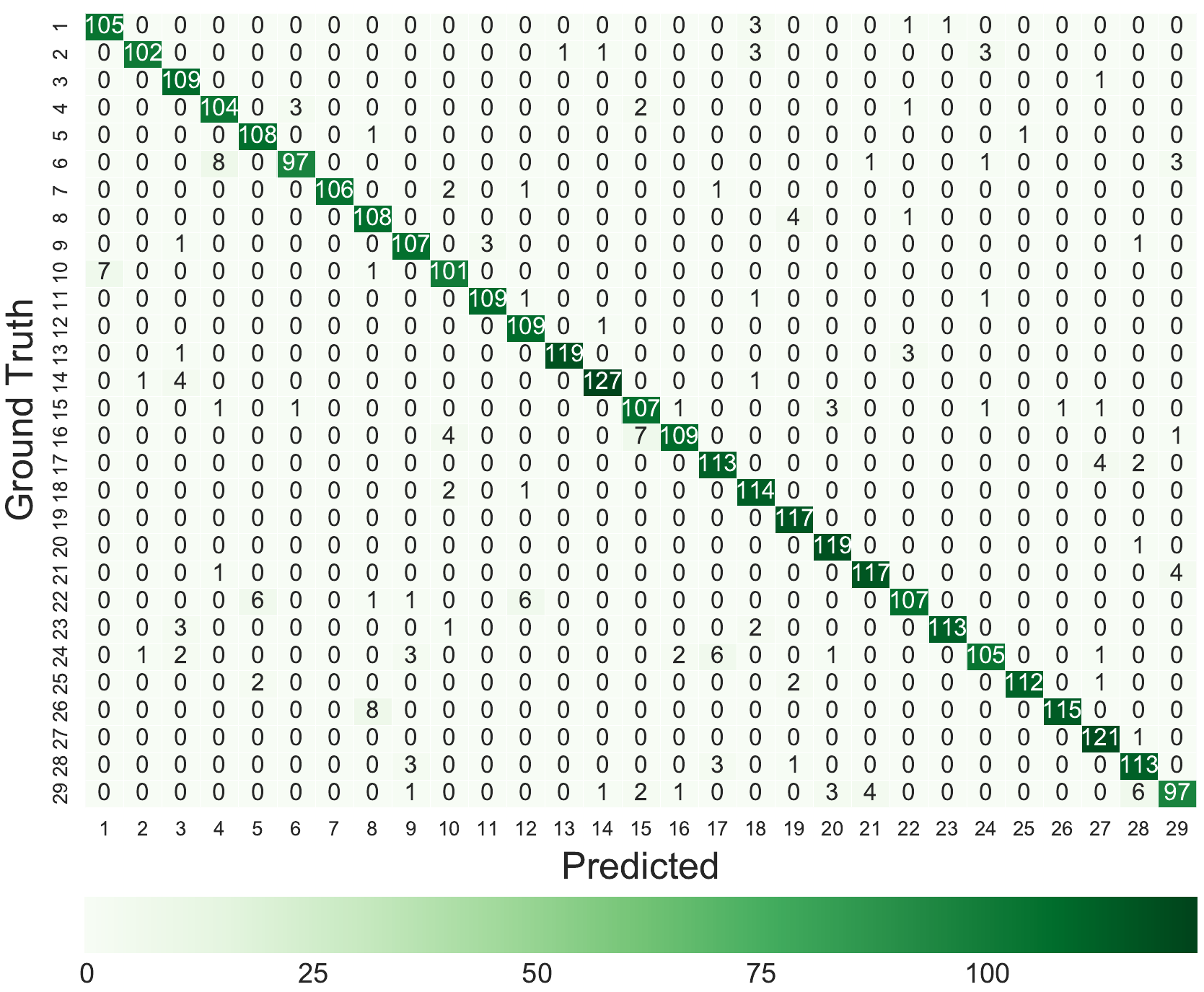}
\vspace{-2mm}
\caption{Confusion matrix of 29 two-hand ASL words.}
\vspace{-3mm}
\label{dia.confusion29}
\end{figure} 

\begin{figure*}[t]
\captionsetup{width=0.31\textwidth}
    \centering
    \begin{minipage}{0.32\textwidth}
        \centering
        \vspace{-1mm}
        \includegraphics[width=0.99\linewidth, keepaspectratio]{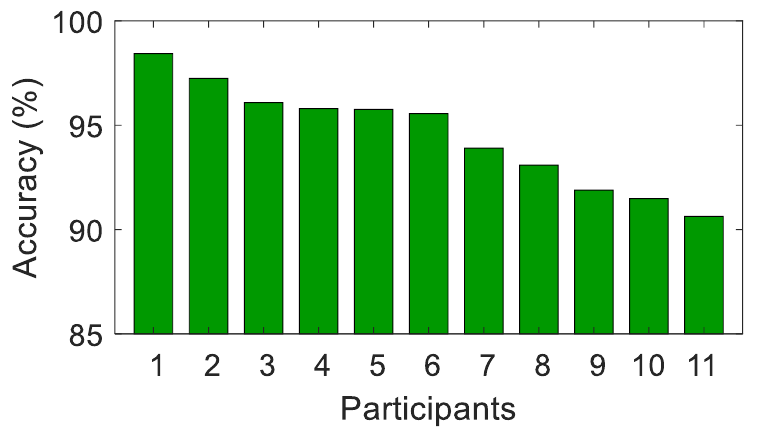}
         
        \caption{Word-level ASL translation accuracy across 11 participants.}
        \vspace{-2mm}
        \label{dia.word_sub}
    \end{minipage}%
    \hspace{3mm}
    \begin{minipage}{0.31\textwidth}
    \vspace{0mm}
        \centering
        \includegraphics[width=0.99\linewidth, keepaspectratio]{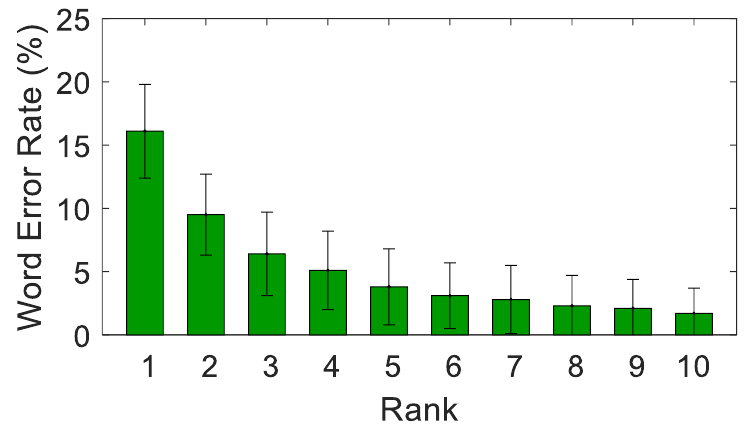}
        \caption{Top-10 $WER$ in translating ASL sentences of unseen participants.}
        \label{dia.sen_rank_sub}
    \end{minipage}
    \hspace{3mm}
    \begin{minipage}{0.31\textwidth}
        \centering
        \vspace{0mm}
        \includegraphics[width=0.99\linewidth, keepaspectratio]{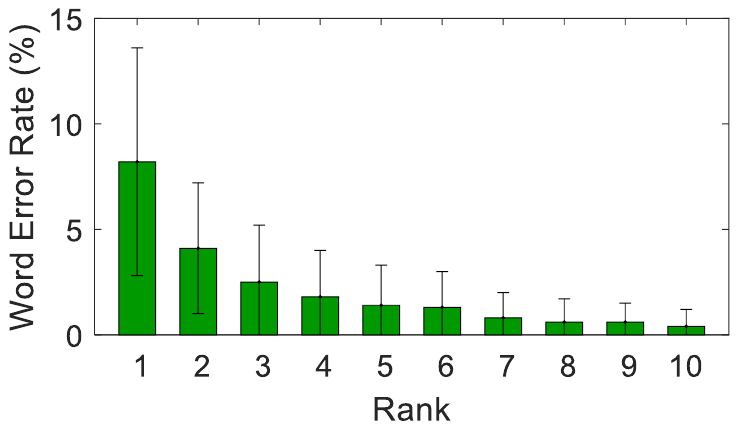}
      
        \caption{Top-10 $WER$ in translating unseen sentences.}
        \label{dia.sen_rank_unseen}
    \end{minipage}
    \vspace{-0mm}
\end{figure*}

\subsection{Word-Level Translation Performance}
%
Figure~\ref{dia.word_sub} shows the average word-level ASL translation accuracy across 11 participants. 
Overall, {\sysname} achieves an average accuracy of $94.5\%$ on translating 56 ASL words across 11 participants. 
This is a very impressive result considering it is achieved based on leave-one-subject-out cross validation protocol. 
Further, we observe that the margin between the highest (participant\#1, 98.4\%) and the lowest (participant\#11, 90.6\%) accuracies is small. 
This indicates that our HB-RNN model is capable of capturing the key characteristics of ASL words. 
Furthermore, the standard deviation of these accuracies is as low as 2.4\%, which also demonstrates the generalization capability of our model across different users. 

To provide a more detailed view of the result, Figure~\ref{dia.confusion27} and~\ref{dia.confusion29} show the confusion matrices of translating 27 one-hand ASL words and 29 two-hand ASL words, respectively. 
As shown in Figure~\ref{dia.confusion27}, among all the 27 one-hand ASL words, only \textit{please} (word\#20) achieves 100\% in both precision and recall. 
This is because \textit{please} has very distinctive hand shape and hand movement characteristics. 
In contrast, \textit{hot} (word\#9) achieves the lowest precision of 81.1\% and \textit{bad} (word\#21) achieves the lowest recall of 86.3\%.
%
Similarly, as shown in Figure~\ref{dia.confusion29}, among all the 29 two-hand ASL words, \textit{big} (word\#7) achieves 100\% in precision and \textit{bicycle} (word\#19) achieves 100\% in recall, whereas \textit{with} (word\#15) has the lowest precision of 90.7\% and \textit{but} (word\#29) has the lowest recall of 84.3\%.
%



\subsection{The Necessity of Model Components}

To validate the design choice of our proposed HB-RNN model, we compare the translation performance between HB-RNN and four comparative models introduced in section 5.2.3.
To make a fair comparison, we set the number of hidden units of each model such that the total number of parameters of each model is roughly the same. 
We use the ASL Words Dataset to evaluate these models. 
%
Table~\ref{tab.compare} lists the evaluation results in terms of average Top-$K$ ($K = 1,2,3$) accuracies and standard deviations across 11 participants.
%
As listed, our proposed HB-RNN model outperforms all four comparative models across all three Top-$K$ metrics. 
Specifically, our HB-RNN model achieves an 5.1\%, 5.0\%, 3.4\% and 0.8\% increase in average Top-$1$ accuracy over the four comparative models, respectively. 
This result demonstrates the superiority of our HB-RNN model over the four comparative models.
%
It also indicates that the hand shape information, hand movement information, the hierarchical structure, as well as the bidirectional connection capture important and complimentary information about the the ASL signs.
By combining these important and complimentary information, the best word-level translation performance is achieved.


\begin{table}[h]
\centering
\scalebox{0.74}{
\begin{tabular}{|c|c|c|c|c|}
\hline
\textbf{Model}     &\textbf{ Top-1 (\%) }   &\textbf{ Top-2 (\%)  }  &\textbf{ Top-3 (\%)  }  & \textbf{Note}      \\ \hline
HB-RNN-M & 89.4 $\pm$ 3.1 & 95.4 $\pm$  1.8 & 97.2 $\pm$ 1.2 & No hand shape    \\ \hline
HB-RNN-S & 89.5 $\pm$ 2.4 & 94.9 $\pm$ 1.7 & 97.0 $\pm$ 1.3 & No hand movement    \\ \hline
SB-RNN   & 91.1 $\pm$ 3.4 & 96.5 $\pm$ 1.7 & 98.2 $\pm$ 1.2 & No hierarchical structure    \\ \hline
H-RNN    & 93.7 $\pm$ 1.7 & 97.1 $\pm$ 0.9 & 98.1 $\pm$ 0.6 & No bidirectional connection   \\ \hline
\textbf{HB-RNN}  &\textbf{ 94.5 $\pm$ 2.4} & \textbf{97.8$\pm$ 1.3}& \textbf{98.7 $\pm$ 0.9}& \textbf{Our model }\\ \hline
\end{tabular}
}
\vspace{2mm}
\caption{Comparison of word-level ASL translation performance between HB-RNN and four comparative models.}
\vspace{-7.5mm}
\label{tab.compare}
\end{table}


\subsection{Sentence-Level Translation Performance}
\vspace{1mm}
\subsubsection{Performance on Unseen Participants}
~\newline
We first evaluate the performance of DeepASL on translating ASL sentences using leave-one-subject-out cross-validation protocol. 
Figure~\ref{dia.sen_rank_sub} shows the results in Top-10 $WER$. 
%
Specifically, the Top-1 $WER$ is 16.1 $\pm$ 3.7\%. 
It indicates that for a 4-word sentence, there is only an average 0.64 words that needs either substitution, deletion or insertion. 
This is a very promising results considering: 1) there are 16 candidate classes (16 ASL words that construct these sentences) in each frame of the sequence; 2) we do not restrict the length or the word order of the sentence during inference and thus there is an enormous amount of possible label sequences; and 3) no language model is leveraged to help improve the performance.


\vspace{1mm}
\subsubsection{Performance on Unseen Sentences}
~\newline
We further conduct an experiment to evaluate the performance of DeepASL on translating unseen sentences (i.e., sentences not included in the training set). 
%
The results are illustrated in Figure~\ref{dia.sen_rank_unseen}.
%
Specifically, the Top-1 $WER$ is 8.2 $\pm$ 5.4\%. 
This indicates that there is only an average 0.33 out of 4 words that needs substitution, deletion and insertion. 
This is a very promising result considering that the translated sentences are not included in the training set.
As a result, it eliminates the burden of collecting all possible ASL sentences.


\subsection{Robustness of ASL Translation in the Field}
Table~\ref{tab.scenario} lists the word-level ASL translation performance achieved on the In-the-Field Dataset.

\vspace{1.5mm}
\noindent
\textbf{Impact of Lighting Conditions:} 
Under poor lighting condition, {\sysname} achieves 96.8 $\pm$ 3.1\% accuracy. 
It indicates that the poor lighting condition has very limited impact on the performance of {\sysname}. 
Under outdoor sunlight condition, {\sysname} achieves 91.8 $\pm$ 4.1\% accuracy. 
This result indicates that the significant portion of infrared light in the sunlight also has very limited impact on the performance of {\sysname}.

\vspace{1.5mm}
\noindent
\textbf{Impact of Body Postures:} {\sysname} achieves 92.2 $\pm$ 3.0\% and 94.9 $\pm$ 4.5\% on walking and standing postures, respectively. The accuracy only drops slightly comparing to previous ASL word recognition result, indicating that {\sysname} could also capture information with either standing or sitting body posture. Moreover, this result also demonstrates the advantage of {\leap} over inertial sensors which are very susceptible to human body motion artifacts.

\vspace{1.5mm}
\noindent
\textbf{Impact of Interference Sources:}  {\sysname} achieves 94.7 $\pm$ 3.0\% and  94.1 $\pm$ 1.3\% on people in-the-scene interference and multi-device interference, respectively. In the first scenario, the accuracy is comparable to previous ASL word recognition result, meaning that {\sysname} is robust to this two interference scenarios. 
We observe that spaced with social distance, {\leap} is rarely confounded by the hands of an interferer. This is because the cameras of {\leap} both have fish-eye angle view, making the far objects too small to be detected. 
As a matter of fact, effective range of {\leap} is designed to be no more than approximately 80 cm~\cite{leapapi}, much less than the social distance. 
On the other hand, our system is not affected by multiple {\leap} present in the ambient environment, indicating that {\sysname} is robust when multiple devices are being used at the same time. 
This is because {\leap} only uses infrared to illuminate the space where ASL is performed and hence the illumination does not have impact on the infrared images captured by the sensor.


\begin{table}[h]
\centering
\vspace{1mm}
\scalebox{0.70}{
\begin{tabular}{|c|c|c|c|c|c|c|}
\hline
\textbf{Category}    		& \multicolumn{2}{c|}{\textbf{Lighting}} & \multicolumn{2}{c|}{\textbf{Body}} & \multicolumn{2}{c|}{\textbf{Interference}} \\ 
\textbf{}    		& \multicolumn{2}{c|}{\textbf{Condition}} & \multicolumn{2}{c|}{\textbf{Posture}} & \multicolumn{2}{c|}{\textbf{Source}} \\ \hline
Subcategory 		& Poor             & Bright            & Walk            & Stand           & People          & Device          \\ \hline
Accuracy (\%)    	&  96.8 $\pm$ 3.1& 91.8 $\pm$ 4.3& 92.2 $\pm$ 3.0& 94.9 $\pm$ 4.5 & 94.7 $\pm$ 3.4 & 94.1 $\pm$ 1.3 \\ \hline
\end{tabular}
}
\vspace{2mm}
\caption{In-the-field ASL translation performance.}
\vspace{-5.5mm}
\label{tab.scenario}
\end{table}



\subsection{System Performance}

To examine the system performance, we have  implemented {\sysname} on three platforms with five computing units: 1) desktop CPU and GPU, 2) mobile CPU and GPU, and 3) tablet CPU. 
%
Our goal is to profile the system performance of {\sysname} across platforms with different computing powers. 
Specifically, we use a desktop installed with an Intel i7-4790 CPU and a Nvidia GTX 1080 GPU to simulate a cloud server; 
we use a mobile development board that contains an ARM Cortex-A57 CPU and a Nvidia Tegra X1 GPU to simulate augmented reality devices with built-in mobile CPU/GPU\footnote{Since Mircrosoft Hololens currently does not support hosting USB clients, we could not implement {\sysname} in  Mircrosoft Hololens to test its system performance.}; 
and we use Microsoft Surface Pro 4 tablet and run {\sysname} on its Intel i5-6300 CPU. 
The specs of the computing units are listed in Table\ref{tab.test_beds}. 



%

To provide a comprehensive evaluation, we evaluate the system performance of three models: 1) one-hand ASL word translation model; 2) two-hand ASL word translation model; and 3) ASL sentence translation model.
In the following, we report their system performance in terms of runtime performance, runtime memory performance, and energy consumption. 

\vspace{-2.5mm}
\begin{table}[h]
	\centering
	\vspace{2mm}
	\scalebox{0.82}{
	\begin{tabular}{|c|c|c|c|c|c|c|}
		\hline
		\multirow{2}{*}{\textbf{Platform}} & \multicolumn{2}{c|}{\textbf{CPU}} & \multirow{2}{*}{\textbf{RAM}} & \multicolumn{3}{c|}{\textbf{GPU}} \\ \cline{2-3} \cline{5-7} 
		& \textbf{Cores} & \textbf{Speed} &  & \textbf{Cores} & \textbf{GFLOPS} & \textbf{Speed} \\ \hline
		Desktop & 8 & 3.6GHz & 16GB & 2560 & 8228 & 1.67GHz \\ \hline
		Mobile  & 4 & 1.9GHz & 4GB & 256 & 512 & 1GHz \\ \hline
		Tablet & 2 & 2.4GHz & 4GB & - & - & \multicolumn{1}{c|}{-} \\ \hline
	\end{tabular}
	}	
	\vspace{2mm}
	\caption{The specs of the three hardware platforms.}
	\label{tab.test_beds}
	\vspace{-7.5mm}
\end{table}


\subsubsection{Runtime Performance}
\begin{figure}[h]
\centering
\vspace{-1mm}
\includegraphics[scale=0.75]{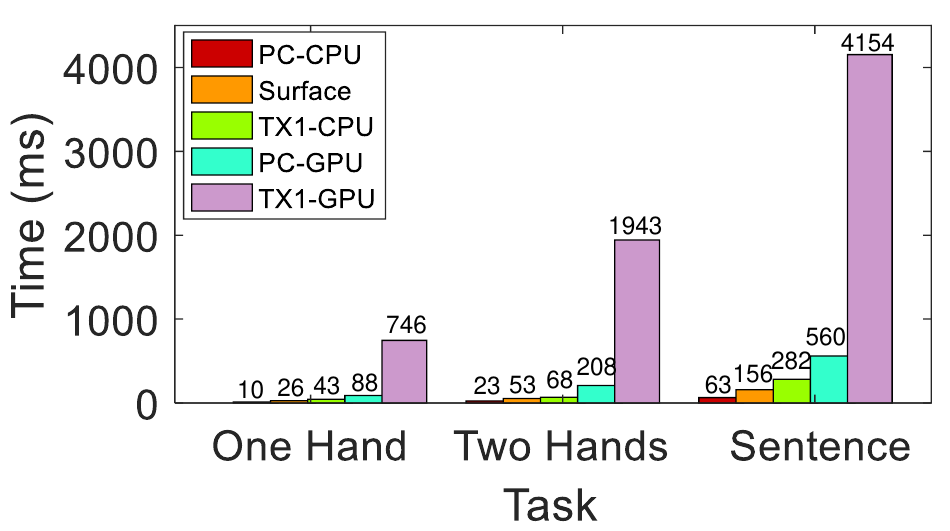}
\vspace{-2mm}
\caption{Runtime performance of the three models.}
\vspace{-2mm}
\label{dia.runtime}
\end{figure}

~\newline
An \textit{inference} contains two parts: data fetching/preprocessing and forward feeding of deep network. Because the time consumed by data fetching/preprocessing is negligible comparing to forward feeding, we report only total \textit{inference} time. 
We run 200 ASL word/sentence recognition and report the average runtime performance. 
The results of runtime performance of three models on five computing units of three platforms are shown in Figure~\ref{dia.runtime}. To give a straightforward view, we order the time cost in an ascending order. 
At a high level, \textit{inference} takes much longer when running on the CPUs than on the GPUs across three models. 
In detail, PC-CPU is 8$\times$ to 9$\times$ faster than PC-GPU; TX1-CPU is 14$\times$ to 28$\times$ faster than TX1-GPU. There are two reasons: (1) during inference, only one sample is passing through the network, which substantially limits the component eligible for parallel computation; and (2) our models are built on RNN, meaning that its time-dependency nature intrinsically eliminate the possibility of parallelism. Therefore, we argue that during inference CPUs are better choice than GPUs. As such, {\sysname} does not need high-end GPU to do inference. 
%
It is also worth pointing out that the runtime of ASL sentence recognition is longer than word recognition. This is because HB-RNN-CTC for sentence recognition has about twice as many parameters as HB-RNN for word recognition. 
Equally important, we observe that the time cost on all three CPUs are less than 282 ms (ASL sentence recognition on TX1-CPU) which means {\sysname} achieves \textit{real-time} recognition performance.

\subsubsection{Runtime Memory Usage}
\begin{figure}[h]
\hspace{1.1mm}
\centering
\includegraphics[scale=0.75]{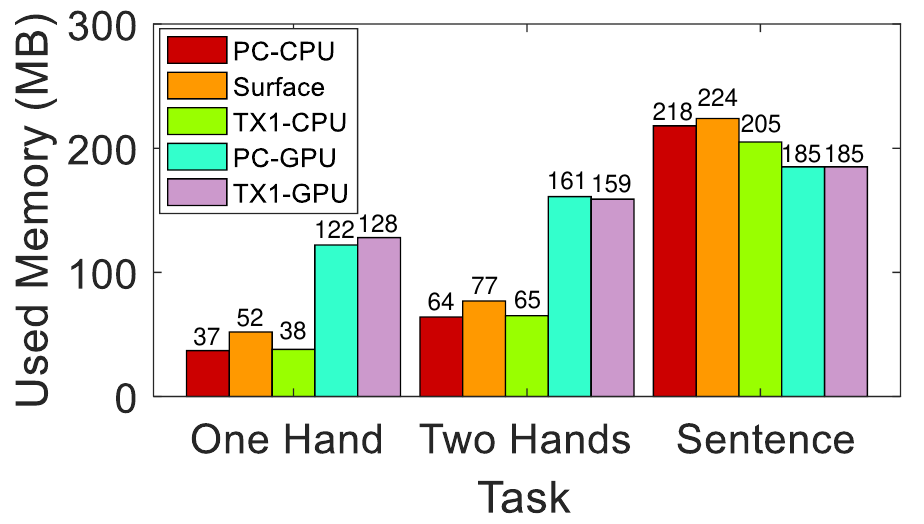}
\vspace{-2mm}
\caption{Runtime memory usage of the three models.}
\vspace{-2mm}
\label{dia.memory}
\end{figure}

~\newline
Next we report the memory cost on various platforms in CPU and GPU mode. Again, we briefly report the average memory usage on three models. Because memory allocation highly depends on operating system and it is difficult to unravel the memory usage for each part, we only report the total memory usage of each model. For all three platforms, we report physical RAM usage and GPU memory usage.
We report these usages because they reflect the memory cost of each model and  might indicate the potential improvements. To clearly reflect the influence of three models on CPU, we report the RAM usage that is subtracted by the RAM usage before doing \textit{inference}. The total RAM usage without loading our model is $2891$ MB on desktop, $931$ MB on TX1 and $1248$ MB on Surface. 
%
%
Figure~\ref{dia.memory} shows the memory cost on five computing units of three platforms. We observe that memory cost of ASL sentence \textit{inference} is larger than two hand ASL word \textit{inference}, which is larger than one hand ASL word. The reason is that in the ASL sentence model, there are more hidden units, thus demanding more allocated memory.

\subsubsection{Energy Consumption}
~\newline
To evaluate the power consumption of {\sysname}, we use PWRcheck DC power analyzer \cite{pwrcheck} to measure the power consumption of both TX1 and Surface tablet. 
We run 200 ASL word/sentence recognition and report the average power consumption.
Table~\ref{tab.consumption} lists the average power consumption of TX1 and Surface respectively. 
We report the power consumption of TX1 to simulate augmented reality devices because the TX1 is designed for mobile device real-time artificial intelligence performance evaluation and thus reflects the portion of power consumed by \textit{inference} in augmented reality devices.
We observe that for TX1 the power consumption of performing inference on GPU is much larger than CPU. This is because: (1) due to the RNN structure of our model, limited amount of computation can be parallelled, making GPU is less efficient in inference than CPU; and (2) performing inference on GPU also involves processing on CPU (for data loading etc.) and thus costs almost twice as much power as CPU alone. 


\begin{table}[h]
\centering
\scalebox{0.83}{
\begin{tabular}{|c|c|c|c|c|}
\hline
\multirow{2}{*}{\textbf{Platform}}                   & \multirow{2}{*}{\textbf{Task}}                 & \textbf{Power} & \textbf{Time} & \textbf{Energy} \\ 
                  &                 & \textbf{(W)} & \textbf{(ms)} & \textbf{(mJ)} \\ \hline
\multirow{7}{*}{\textbf{TX1}}     & Idle                  & 3.54     & -         & -           \\ \cline{2-5} 
                         & One-hand ASL word (CPU)   & 5.92     & 42.9       & 254.0      \\ \cline{2-5} 
                         & Two-hand ASL word (CPU) & 6.13     & 66.1     & 417.5      \\ \cline{2-5} 
                         & ASL sentence (CPU)        & 6.02     & 281.7     & 1695.8      \\ \cline{2-5} 
                         & One-hand ASL word (GPU)   & 12.31    & 746.2       &9185.7     \\ \cline{2-5} 
                         & Two-hand ASL word (GPU)  & 12.16    & 1943.4     &23631.7     \\ \cline{2-5} 
                         & ASL sentence (GPU)        & 11.75    & 4153.6   & 48804.8     \\ \hline
\multirow{6}{*}{\textbf{Surface}} & Sleep                 & 1.63     & -         & -           \\ \cline{2-5} 
                         & Screen-on             & 8.32     & -         & -           \\ \cline{2-5} 
                         & ASL Dictionary App-on     & 15.75    & -         & -           \\ \cline{2-5} 
                         & One-hand ASL word         & 23.67    & 26.1      & 591.7      \\ \cline{2-5} 
                         & Two-hand ASL word        & 24.21    & 52.7       &1117.8      \\ \cline{2-5} 
                         & ASL sentence              & 22.13    & 156.2     & 3456.7      \\ \hline
\end{tabular}
}
\vspace{2mm}
\caption{Energy consumption on TX1 and Surface.}
\vspace{-6.5mm}
\label{tab.consumption}
\end{table}

Finally, in Table~\ref{tab.num_infer}, we report the estimated number of ASL word/sentence recognition that can be completed by TX1 and Surface, using fully-charged battery of Hololens (16.5 Wh) and Surface (38.2 Wh), respectively.
For TX1, the number of inferences of CPU is 36$\times$, 57$\times$ and 29$\times$ larger than those of its GPU for three model respetively. It means that in terms of performing \textit{inference}, CPU is more suitable. Meanwhile, despite the power consumption from other sources, a Hololens/Surface equal volume battery could support enough number of inferences within one day.

\begin{table}[h]
\centering
\scalebox{0.83}{
\begin{tabular}{|c|c|c|c|}
\hline
\textbf{Platform}                 & \textbf{Task}           & \textbf{CPU}    & \textbf{GPU}  \\ \hline
\multirow{3}{*}{\textbf{TX1}}     & One-hand ASL word & 233888 & 6467 \\ \cline{2-4} 
                         & Two-hand ASL word & 142291  & 2514 \\ \cline{2-4} 
                         & ASL sentence       & 35028  & 1217 \\ \hline
\multirow{3}{*}{\textbf{Surface}} & One-hand ASL word  & 232420  & -    \\ \cline{2-4} 
                         & Two-hand ASL word & 123031  & -    \\ \cline{2-4} 
                         & ASL sentence       & 39784  & -    \\ \hline
\end{tabular}
}
\vspace{2mm}
\caption{Estimated number of inferences on TX1 and Surface with a 16.5 Wh (Hololens) and 38.2 Wh (Surface) battery, respectively.}
\vspace{-7.5mm}
\label{tab.num_infer}
\end{table}


\section{Applications}

The design of {\sysname} enables a wide range of applications. 
To demonstrate the practical value of {\sysname}, we have developed two prototype applications based on {\sysname}.
In this section, we briefly describe these two prototype applications.

\subsection{Application\#1: Personal Interpreter}

\vspace{0.5mm}
\noindent
\textbf{Use Scenario:} 
For the first application, {\sysname} is used as a \textit{Personal Interpreter}.
%
With the help of an AR headset, \textit{Personal Interpreter} enables real-time two-way communications between a deaf person and peole who do not understand ASL.
Specifically, on one hand, \textit{Personal Interpreter} uses speech recognition technology to translate spoken languages into digital texts, and then projects the digital texts to the AR screen for the deaf person to see; 
on the other hand, \textit{Personal Interpreter} uses ASL recognition technology to translate ASL performed by the deaf person into spoken languages for peole who do not understand ASL.

\begin{figure}[h]
\centering
\includegraphics[scale=0.2]{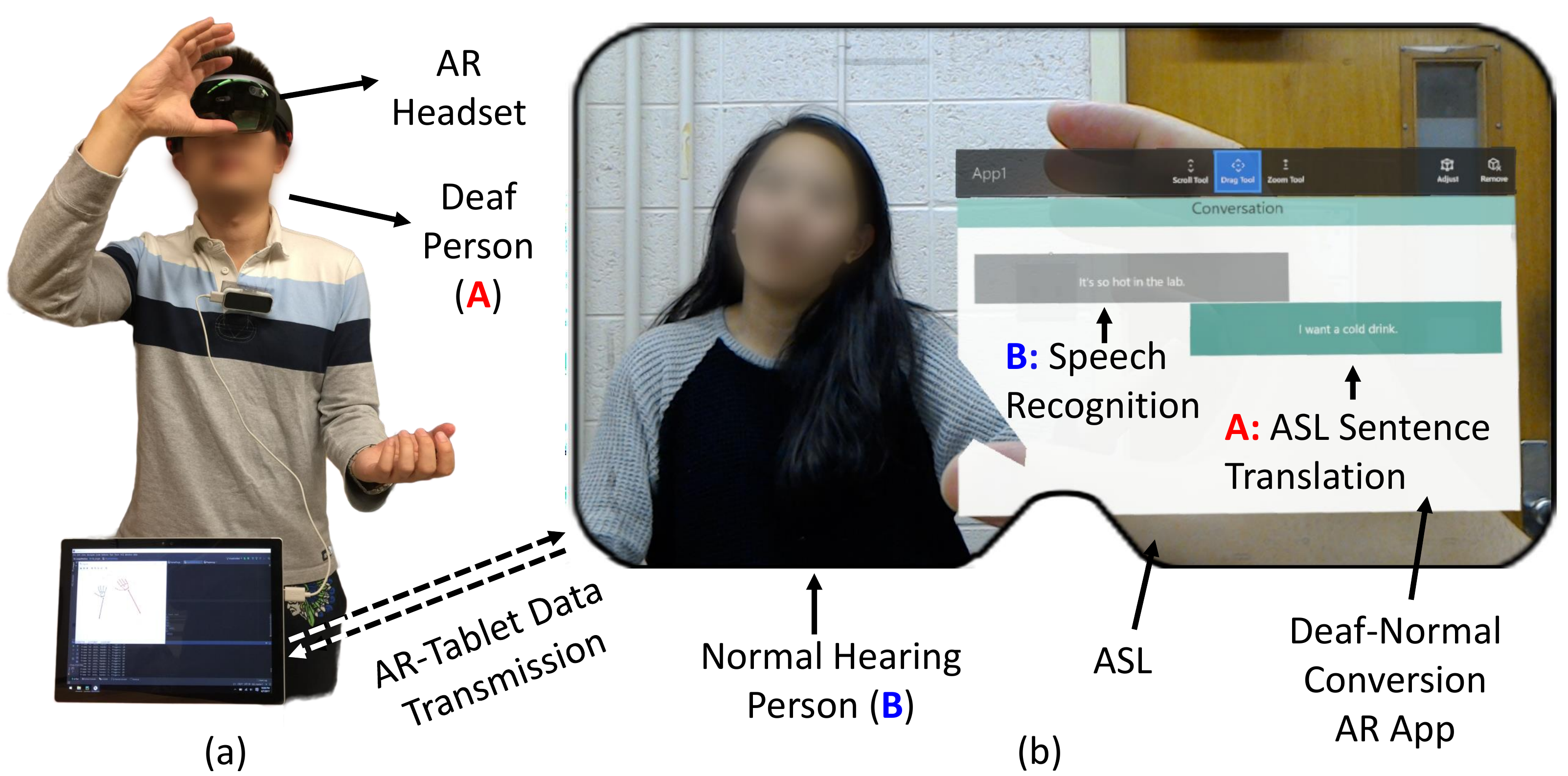}
\caption{The \textit{Personal Interpreter} application: (a) a deaf person performing ASL while wearing a Microsoft Hololens AR headset; (b) a screenshot from AR perspective.}
\vspace{-2mm}
\label{dia.holo}
\end{figure}

\vspace{0.5mm}
\noindent
\textbf{Implementation:}
We implemented \textit{Personal Interpreter} as a Microsoft Hololens application. 
Since Mircrosoft Hololens currently does not support hosting USB clients, we could not implement DeepASL in Mircrosoft Hololens.
Instead, we transmitted the ASL recognition results to Hololens via TCP/IP.
Figure~\ref{dia.holo} illustrates the usage scenario and a screenshot from AR perspective. 
As shown, the recognized ASL sentence is displayed in the green dialogue box in the Hololens application. 
%
%
The tablet-AR set is burdensome to the deaf people, but we envision that in the future, the AR headset will be miniaturized and hence is much less burdensome for people to wear on a daily basis. 
Meanwhile, the future AR headset will be able to host a USB device, enabling direct data transmission from {\leap} to itself.



\subsection{Application\#2: ASL Dictionary}

\vspace{0.5mm}
\noindent
\textbf{Use Scenario:} 
For the second application, {\sysname} is used as an \textit{ASL Dictionary} to help a deaf person look up unknown ASL words.
%
%
Spoken languages (e.g., English) allow one to easily look up an unknown word via indexing. 
Unfortunately, this is not the case for ASL. 
Imagine a deaf child who wants to look up an ASL word that she remembers how to perform but forgets the meaning of it. 
Without the help of a person who understands ASL, there is not an easy way for her to look up the ASL word. 
This is because unlike spoken languages, ASL does not have a natural form to properly index each gesture. 
\textit{ASL Dictionary} solves this problem by taking the sign of the ASL word as input and displays the meaning of this ASL word in real time. 

\begin{figure}[h]
\vspace{0.2mm}
\centering
\includegraphics[scale=0.18]{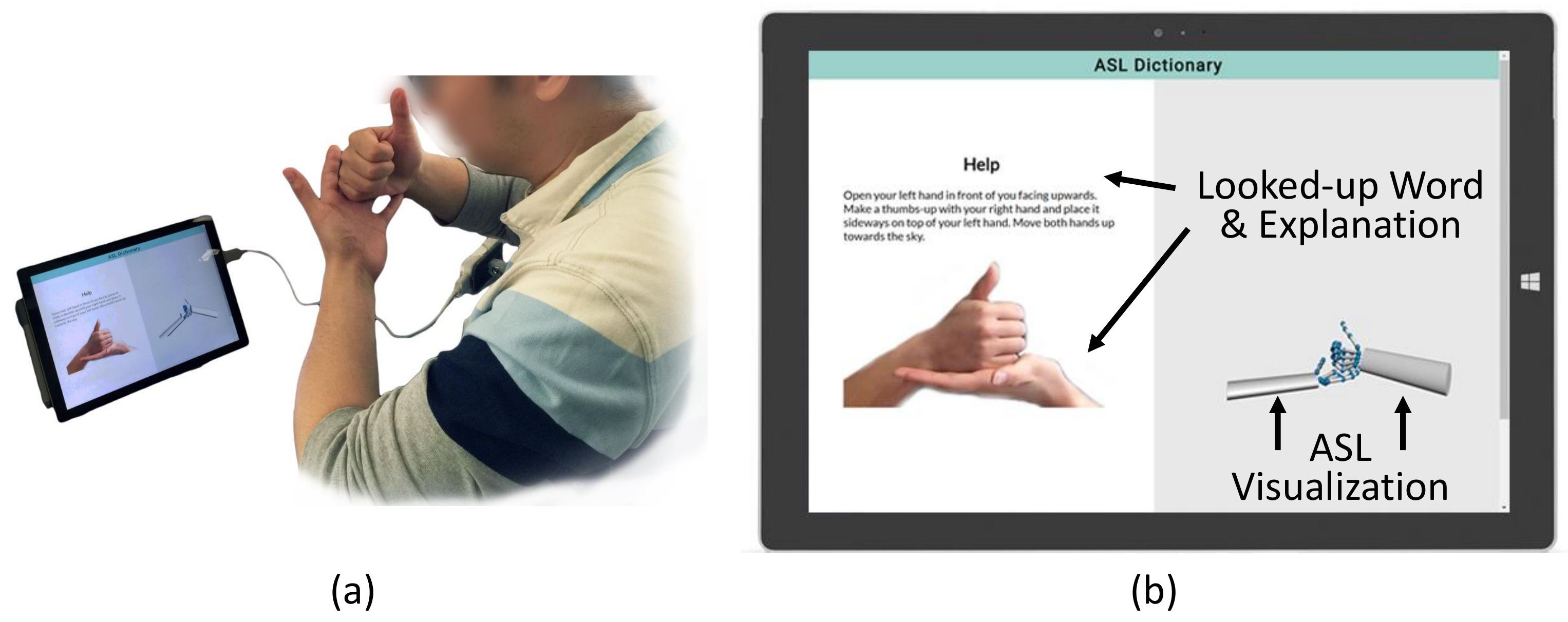}
\vspace{-2mm}
\caption{The \textit{ASL Dictionary} application: (a) a deaf person looking up an ASL word ``help''; (b) a screenshot of the tablet application.}
\vspace{-2mm}
\label{dia.dict}
\end{figure}

\vspace{0.5mm}
\noindent
\textbf{Implementation:} 
We implemented \textit{ASL Dictionary} as a Microsoft Surface tablet application.
Figure~\ref{dia.dict} illustrates the usage scenario and a screenshot of the tablet application.
%

\vspace{-2mm}

\section{Discussion}
\label{sec.diss}

%


\vspace{1mm}

\noindent
\textbf{Impact on ASL Translation Technology:} 
{\sysname} represents the first ASL translation technology that enables ubiquitous and non-intrusive ASL translation at both word and sentence levels.
It demonstrates the superiority of using infrared light and skeleton joints information over other sensing modalities for capturing key characteristics of ASL signs.
It also demonstrates the capability of hierarchical bidirectional deep recurrent neural network for single-sign modeling as well as CTC for translating the whole sentence end-to-end without requiring users to pause between adjacent signs.
Given the innovation solution it provides and its promising performance, we believe {\sysname} has made a significant contribution to the advancement of ASL translation technology.
%
%



\vspace{1mm}

\noindent
\textbf{Initiative on ASL Sign Data Crowdsourcing:} 
%
Despite months of efforts spent on data collection, our dataset still covers a small portion of the ASL vocabulary.
To make {\sysname} being able to translate as many words as in the ASL vocabulary and many more sentences that deaf people use in their daily-life communications, we have taken an initiative on ASL sign data crowdsourcing. 
We have made our data collection toolkit publicly available.
%
We hope our initiative could serve as a seed to draw attentions from people who share the same vision as we have, and want to turn this vision into reality. 
We deeply believe that, with the crowdsourced efforts, ASL translation technology can be significantly advanced.
With that, we could ultimately turn the vision of tearing down the communication barrier between the deaf people and the hearing majority into reality.



%
%



%
%

\vspace{-1mm}

\section{Related Work}
\label{sec.rw}

\noindent
%
Our work is related to two research areas: 1) sign language translation; and more broadly 2) mobile sensing systems. 

\vspace{1mm}
\noindent
\textbf{Sign Language Translation Systems:}
Over the past few decades, a variety of sign language translation systems based on various sensing modalities have been developed.
%
Among them, systems based on wearable sensors have been extensively studied~\cite{kim2008bi, kosmidou2009sign, li2010automatic, li2012sign, wu2015real, tubaiz2015glove, praveen2014sign, abhishek2016glove, kanwal2014assistive}. 
These systems use motion sensors, EMG sensors, bend sensors, or their combinations to capture hand movements, muscle activities, or bending of fingers to infer the performed signs.
For example, Wu \textit{et al}. developed a wrist-worn device with onboard motion and EMG sensors which is able to recognize $40$ signs~\cite{wu2015real}. 
%
%
%
Another widely used sensing modality in sign language translation systems is RGB camera~\cite{starner1998real,zafrulla2010novel,brashear2003using}.
%
For example, Starner \textit{et al}. are able to recognize 40 ASL words using Hidden Markov Model with a hat-mounted RGB camera~\cite{starner1998real}.
%
%
There are also some efforts on designing sign language translation systems based on Microsoft Kinect~\cite{chai2013sign, chai2013visualcomm}.
%
As an example, by capturing the skeleton joints of the user body and limbs using Microsoft Kinect, Chai \textit{et al}. are able to recognize Chinese Sign Language by matching the collected skeleton trajectory with gallery trajectories~\cite{chai2013sign}.
Most recently, researchers have started exploring using {\leap} to build sign language translation systems~\cite{chuan2014american,marin2014hand}. 
%
However, these systems are very limited in their capabilities in the sense that they can only recognize static ASL signs by capturing hand shape information.
%
In contrast, {\sysname} captures both hand shape and movement information so that it is able to recognize dynamic signs that involve movements.
Most importantly, compared to all the existing sign language translation systems, {\sysname} is the first framework that enables end-to-end ASL sentence translation.

\vspace{1mm}
\noindent
\textbf{Mobile Sensing Systems:}
%
Our work is also broadly related to research in mobile sensing systems.
Prior mobile sensing systems have explored a variety of sensing modalities that have enabled a wide range of innovative applications.
Among them, accelerometer, microphone and physiological sensors are some of the mostly explored sensing modalities.
For example, Mokaya \textit{et al.} developed an accelerometer-based system to sense skeletal muscle vibrations for quantifying skeletal muscle fatigue in an exercise setting \cite{mokaya2016burnout}. 
%
Nirjon \textit{et al.} developed MusicalHeart \cite{nirjon2012musicalheart} which integrated a microphone into an earphone to extract heartbeat information from audio signals.
Nguyen \textit{et al.} designed an in-ear sensing system in the form of earplugs that is able to capture EEG, EOG, and EMG signals for sleep monitoring \cite{nguyen2016lightweight}.
Recently, researchers have started exploring using wireless radio signal as a contactless sensing mechanism. 
For example, Wang \textit{et al.} developed WiFall \cite{wang2017wifall} that used wireless radio signal to detect accidental falls.
Fang \textit{et al.} used radio as a single sensing modality for integrated activities of daily living and vital sign monitoring \cite{fang2016bodyscan}.
In this work, we explore infrared light as a new sensing modality in the context of ASL translation.
It complements existing mobile sensing systems by providing a non-intrusive and high-resolution sensing scheme.
%
We regard this work as an excellent example to demonstrate the usefulness of infrared sensing for mobile systems.
With the incoming era of virtual/augmented reality, we envision infrared sensing will be integrated into many future mobile systems such as smartphones and smart glasses.


\vspace{-2mm}

\section{Conclusion}
\label{sec.con}
\noindent
In this paper, we present the design, implementation and evaluation of {\sysname}, a transformative deep learning-based sign language translation technology that enables ubiquitous and non-intrusive ASL translation at both word and sentence levels.
%
%
At the word level, {\sysname} achieves an average $94.5\%$ translation accuracy over 56 commonly used ASL words. 
At the sentence level, {\sysname} achieves an average $8.2\%$ word error rate on translating unseen ASL sentences and an average $16.1\%$ word error rate on translating ASL sentences performed by unseen users over 100 commonly used ASL sentences. 
%
%
%
Given the innovation solution it provides and its promising performance, we believe DeepASL has made a significant contribution to the advancement of ASL translation technology.

\begin{acks}

\noindent
We would like to thank Dr. John Stankovic for being the shepherd of this paper. 
We are also grateful to the anonymous SenSys reviewers for their valuable reviews and insightful comments. 
This research was partially funded by NSF awards \#1565604 and \#1617627.
\end{acks}

\balance
\bibliographystyle{ACM-Reference-Format}
\bibliography{SenSys2017} 


\begin{thebibliography}{00}


\ifx \showCODEN    \undefined \def \showCODEN     #1{\unskip}     \fi
\ifx \showDOI      \undefined \def \showDOI       #1{#1}\fi
\ifx \showISBNx    \undefined \def \showISBNx     #1{\unskip}     \fi
\ifx \showISBNxiii \undefined \def \showISBNxiii  #1{\unskip}     \fi
\ifx \showISSN     \undefined \def \showISSN      #1{\unskip}     \fi
\ifx \showLCCN     \undefined \def \showLCCN      #1{\unskip}     \fi
\ifx \shownote     \undefined \def \shownote      #1{#1}          \fi
\ifx \showarticletitle \undefined \def \showarticletitle #1{#1}   \fi
\ifx \showURL      \undefined \def \showURL       {\relax}        \fi
\providecommand\bibfield[2]{#2}
\providecommand\bibinfo[2]{#2}
\providecommand\natexlab[1]{#1}
\providecommand\showeprint[2][]{arXiv:#2}

\bibitem[\protect\citeauthoryear{??}{pwr}{2016}]%
        {pwrcheck}
 \bibinfo{year}{2016}\natexlab{}.
\newblock \bibinfo{title}{{PWRcheck DC power analyzer}}.
\newblock
  \bibinfo{howpublished}{\url{http://www.westmountainradio.com/product_info.php?products_id=pwrcheck}}.
    (\bibinfo{year}{2016}).
\newblock


\bibitem[\protect\citeauthoryear{??}{dea}{2017a}]%
        {deafstats1}
 \bibinfo{year}{2017}\natexlab{a}.
\newblock \bibinfo{title}{{American Deaf And Hard of Hearing Statistics}}.
\newblock
  \bibinfo{howpublished}{\url{http://www.ncra.org/Government/content.cfm?ItemNumber=9450}}.
    (\bibinfo{year}{2017}).
\newblock


\bibitem[\protect\citeauthoryear{??}{dea}{2017b}]%
        {deafstats2}
 \bibinfo{year}{2017}\natexlab{b}.
\newblock \bibinfo{title}{{American Sign Language | NIDCD}}.
\newblock
  \bibinfo{howpublished}{\url{https://www.nidcd.nih.gov/health/american-sign-language}}.
    (\bibinfo{year}{2017}).
\newblock


\bibitem[\protect\citeauthoryear{??}{lea}{2017a}]%
        {leapmotion}
 \bibinfo{year}{2017}\natexlab{a}.
\newblock \bibinfo{title}{{Leap Motion}}.
\newblock \bibinfo{howpublished}{\url{https://www.leapmotion.com/}}.
  (\bibinfo{year}{2017}).
\newblock


\bibitem[\protect\citeauthoryear{??}{lea}{2017b}]%
        {leapapi}
 \bibinfo{year}{2017}\natexlab{b}.
\newblock \bibinfo{title}{Leap Motion API}.
\newblock   (\bibinfo{year}{2017}).
\newblock
\newblock
\shownote{\url{https://developer.leapmotion.com/documentation/python/api/Leap_Classes.html}.}


\bibitem[\protect\citeauthoryear{??}{who}{2017}]%
        {who}
 \bibinfo{year}{2017}\natexlab{}.
\newblock \bibinfo{title}{{WHO | Deafness and hearing loss}}.
\newblock
  \bibinfo{howpublished}{\url{http://www.who.int/mediacentre/factsheets/fs300/en/}}.
    (\bibinfo{year}{2017}).
\newblock


\bibitem[\protect\citeauthoryear{??}{sig}{2017}]%
        {signlanguagelist}
 \bibinfo{year}{2017}\natexlab{}.
\newblock \bibinfo{title}{{Wikipedia | List of sign languages}}.
\newblock
  \bibinfo{howpublished}{\url{https://en.wikipedia.org/wiki/List_of_sign_languages}}.
    (\bibinfo{year}{2017}).
\newblock


\bibitem[\protect\citeauthoryear{Abhishek, Qubeley, and Ho}{Abhishek
  et~al\mbox{.}}{2016}]%
        {abhishek2016glove}
\bibfield{author}{\bibinfo{person}{Kalpattu~S Abhishek}, \bibinfo{person}{Lee
  Chun~Fai Qubeley}, {and} \bibinfo{person}{Derek Ho}.}
  \bibinfo{year}{2016}\natexlab{}.
\newblock \showarticletitle{Glove-based hand gesture recognition sign language
  translator using capacitive touch sensor}. In \bibinfo{booktitle}{{\em
  Electron Devices and Solid-State Circuits (EDSSC), 2016 IEEE International
  Conference on}}. IEEE, \bibinfo{pages}{334--337}.
\newblock


\bibitem[\protect\citeauthoryear{Assael, Shillingford, Whiteson, and
  de~Freitas}{Assael et~al\mbox{.}}{2016}]%
        {assael2016lipnet}
\bibfield{author}{\bibinfo{person}{Yannis~M Assael}, \bibinfo{person}{Brendan
  Shillingford}, \bibinfo{person}{Shimon Whiteson}, {and}
  \bibinfo{person}{Nando de Freitas}.} \bibinfo{year}{2016}\natexlab{}.
\newblock \showarticletitle{LipNet: Sentence-level Lipreading}.
\newblock \bibinfo{journal}{{\em arXiv preprint arXiv:1611.01599\/}}
  (\bibinfo{year}{2016}).
\newblock


\bibitem[\protect\citeauthoryear{Brashear, Starner, Lukowicz, and
  Junker}{Brashear et~al\mbox{.}}{2003}]%
        {brashear2003using}
\bibfield{author}{\bibinfo{person}{Helene Brashear}, \bibinfo{person}{Thad
  Starner}, \bibinfo{person}{Paul Lukowicz}, {and} \bibinfo{person}{Holger
  Junker}.} \bibinfo{year}{2003}\natexlab{}.
\newblock \showarticletitle{Using Multiple Sensors for Mobile Sign Language
  Recognition}. In \bibinfo{booktitle}{{\em Proceedings of the 7th IEEE
  International Symposium on Wearable Computers}} {\em (\bibinfo{series}{ISWC
  '03})}. \bibinfo{publisher}{IEEE Computer Society},
  \bibinfo{address}{Washington, DC, USA}, \bibinfo{pages}{45--}.
\newblock
\showISBNx{0-7695-2034-0}
\showURL{%
\url{http://dl.acm.org/citation.cfm?id=946249.946868}}


\bibitem[\protect\citeauthoryear{Chai, Li, Chen, Zhou, Wu, and Li}{Chai
  et~al\mbox{.}}{2013a}]%
        {chai2013visualcomm}
\bibfield{author}{\bibinfo{person}{Xiujuan Chai}, \bibinfo{person}{Guang Li},
  \bibinfo{person}{Xilin Chen}, \bibinfo{person}{Ming Zhou},
  \bibinfo{person}{Guobin Wu}, {and} \bibinfo{person}{Hanjing Li}.}
  \bibinfo{year}{2013}\natexlab{a}.
\newblock \showarticletitle{Visualcomm: A tool to support communication between
  deaf and hearing persons with the kinect}. In \bibinfo{booktitle}{{\em
  Proceedings of the 15th International ACM SIGACCESS Conference on Computers
  and Accessibility}}. ACM, \bibinfo{pages}{76}.
\newblock


\bibitem[\protect\citeauthoryear{Chai, Li, Lin, Xu, Tang, Chen, and Zhou}{Chai
  et~al\mbox{.}}{2013b}]%
        {chai2013sign}
\bibfield{author}{\bibinfo{person}{Xiujuan Chai}, \bibinfo{person}{Guang Li},
  \bibinfo{person}{Yushun Lin}, \bibinfo{person}{Zhihao Xu},
  \bibinfo{person}{Yili Tang}, \bibinfo{person}{Xilin Chen}, {and}
  \bibinfo{person}{Ming Zhou}.} \bibinfo{year}{2013}\natexlab{b}.
\newblock \showarticletitle{Sign language recognition and translation with
  kinect}. In \bibinfo{booktitle}{{\em IEEE Conf. on AFGR}}.
\newblock


\bibitem[\protect\citeauthoryear{Chuan, Regina, and Guardino}{Chuan
  et~al\mbox{.}}{2014}]%
        {chuan2014american}
\bibfield{author}{\bibinfo{person}{Ching-Hua Chuan}, \bibinfo{person}{Eric
  Regina}, {and} \bibinfo{person}{Caroline Guardino}.}
  \bibinfo{year}{2014}\natexlab{}.
\newblock \showarticletitle{American Sign Language recognition using leap
  motion sensor}. In \bibinfo{booktitle}{{\em Machine Learning and Applications
  (ICMLA), 2014 13th International Conference on}}. IEEE,
  \bibinfo{pages}{541--544}.
\newblock


\bibitem[\protect\citeauthoryear{Dominio, Donadeo, and Zanuttigh}{Dominio
  et~al\mbox{.}}{2014}]%
        {dominio2014combining}
\bibfield{author}{\bibinfo{person}{Fabio Dominio}, \bibinfo{person}{Mauro
  Donadeo}, {and} \bibinfo{person}{Pietro Zanuttigh}.}
  \bibinfo{year}{2014}\natexlab{}.
\newblock \showarticletitle{Combining multiple depth-based descriptors for hand
  gesture recognition}.
\newblock \bibinfo{journal}{{\em Pattern Recognition Letters\/}}
  \bibinfo{volume}{50} (\bibinfo{year}{2014}), \bibinfo{pages}{101--111}.
\newblock


\bibitem[\protect\citeauthoryear{Donahue, Anne~Hendricks, Guadarrama, Rohrbach,
  Venugopalan, Saenko, and Darrell}{Donahue et~al\mbox{.}}{2015}]%
        {donahue2015long}
\bibfield{author}{\bibinfo{person}{Jeffrey Donahue}, \bibinfo{person}{Lisa
  Anne~Hendricks}, \bibinfo{person}{Sergio Guadarrama}, \bibinfo{person}{Marcus
  Rohrbach}, \bibinfo{person}{Subhashini Venugopalan}, \bibinfo{person}{Kate
  Saenko}, {and} \bibinfo{person}{Trevor Darrell}.}
  \bibinfo{year}{2015}\natexlab{}.
\newblock \showarticletitle{Long-term recurrent convolutional networks for
  visual recognition and description}. In \bibinfo{booktitle}{{\em Proceedings
  of the IEEE conference on computer vision and pattern recognition}}.
  \bibinfo{pages}{2625--2634}.
\newblock


\bibitem[\protect\citeauthoryear{Du, Wang, and Wang}{Du et~al\mbox{.}}{2015}]%
        {du2015hierarchical}
\bibfield{author}{\bibinfo{person}{Yong Du}, \bibinfo{person}{Wei Wang}, {and}
  \bibinfo{person}{Liang Wang}.} \bibinfo{year}{2015}\natexlab{}.
\newblock \showarticletitle{Hierarchical recurrent neural network for skeleton
  based action recognition}. In \bibinfo{booktitle}{{\em Proceedings of the
  IEEE conference on computer vision and pattern recognition}}.
  \bibinfo{pages}{1110--1118}.
\newblock


\bibitem[\protect\citeauthoryear{Fang, Lane, Zhang, Boran, and Kawsar}{Fang
  et~al\mbox{.}}{2016}]%
        {fang2016bodyscan}
\bibfield{author}{\bibinfo{person}{Biyi Fang}, \bibinfo{person}{Nicholas~D.
  Lane}, \bibinfo{person}{Mi Zhang}, \bibinfo{person}{Aidan Boran}, {and}
  \bibinfo{person}{Fahim Kawsar}.} \bibinfo{year}{2016}\natexlab{}.
\newblock \showarticletitle{BodyScan: Enabling Radio-based Sensing on Wearable
  Devices for Contactless Activity and Vital Sign Monitoring}. In
  \bibinfo{booktitle}{{\em The 14th ACM International Conference on Mobile
  Systems, Applications, and Services (MobiSys)}}. \bibinfo{pages}{97--110}.
\newblock


\bibitem[\protect\citeauthoryear{Graves et~al\mbox{.}}{Graves
  et~al\mbox{.}}{2012}]%
        {graves2012supervised}
\bibfield{author}{\bibinfo{person}{Alex Graves} {et~al\mbox{.}}}
  \bibinfo{year}{2012}\natexlab{}.
\newblock \bibinfo{booktitle}{{\em Supervised sequence labelling with recurrent
  neural networks}}. Vol.~\bibinfo{volume}{385}.
\newblock \bibinfo{publisher}{Springer}.
\newblock


\bibitem[\protect\citeauthoryear{Graves, Fern{\'a}ndez, Gomez, and
  Schmidhuber}{Graves et~al\mbox{.}}{2006}]%
        {graves2006connectionist}
\bibfield{author}{\bibinfo{person}{Alex Graves}, \bibinfo{person}{Santiago
  Fern{\'a}ndez}, \bibinfo{person}{Faustino Gomez}, {and}
  \bibinfo{person}{J{\"u}rgen Schmidhuber}.} \bibinfo{year}{2006}\natexlab{}.
\newblock \showarticletitle{Connectionist temporal classification: labelling
  unsegmented sequence data with recurrent neural networks}. In
  \bibinfo{booktitle}{{\em Proceedings of the 23rd international conference on
  Machine learning}}. ACM, \bibinfo{pages}{369--376}.
\newblock


\bibitem[\protect\citeauthoryear{Graves, Mohamed, and Hinton}{Graves
  et~al\mbox{.}}{2013}]%
        {graves2013speech}
\bibfield{author}{\bibinfo{person}{Alex Graves}, \bibinfo{person}{Abdel-rahman
  Mohamed}, {and} \bibinfo{person}{Geoffrey Hinton}.}
  \bibinfo{year}{2013}\natexlab{}.
\newblock \showarticletitle{Speech recognition with deep recurrent neural
  networks}. In \bibinfo{booktitle}{{\em Acoustics, speech and signal
  processing (icassp), 2013 ieee international conference on}}. IEEE,
  \bibinfo{pages}{6645--6649}.
\newblock


\bibitem[\protect\citeauthoryear{Hochreiter, Bengio, Frasconi, and
  Schmidhuber}{Hochreiter et~al\mbox{.}}{2001}]%
        {hochreiter2001gradient}
\bibfield{author}{\bibinfo{person}{Sepp Hochreiter}, \bibinfo{person}{Yoshua
  Bengio}, \bibinfo{person}{Paolo Frasconi}, {and} \bibinfo{person}{J{\"u}rgen
  Schmidhuber}.} \bibinfo{year}{2001}\natexlab{}.
\newblock \bibinfo{title}{Gradient flow in recurrent nets: the difficulty of
  learning long-term dependencies}.
\newblock   (\bibinfo{year}{2001}).
\newblock


\bibitem[\protect\citeauthoryear{Hochreiter and Schmidhuber}{Hochreiter and
  Schmidhuber}{1997}]%
        {hochreiter1997long}
\bibfield{author}{\bibinfo{person}{Sepp Hochreiter} {and}
  \bibinfo{person}{J{\"u}rgen Schmidhuber}.} \bibinfo{year}{1997}\natexlab{}.
\newblock \showarticletitle{Long short-term memory}.
\newblock \bibinfo{journal}{{\em Neural computation\/}} \bibinfo{volume}{9},
  \bibinfo{number}{8} (\bibinfo{year}{1997}), \bibinfo{pages}{1735--1780}.
\newblock


\bibitem[\protect\citeauthoryear{Hoza}{Hoza}{2007}]%
        {hoza2007s}
\bibfield{author}{\bibinfo{person}{Jack Hoza}.}
  \bibinfo{year}{2007}\natexlab{}.
\newblock \bibinfo{booktitle}{{\em It's not what you sign, it's how you sign
  it: politeness in American Sign Language}}.
\newblock \bibinfo{publisher}{Gallaudet University Press}.
\newblock


\bibitem[\protect\citeauthoryear{Kanwal, Abdullah, Ahmed, Saher, and
  Jafri}{Kanwal et~al\mbox{.}}{2014}]%
        {kanwal2014assistive}
\bibfield{author}{\bibinfo{person}{Kehkashan Kanwal}, \bibinfo{person}{Saad
  Abdullah}, \bibinfo{person}{Yusra~Binte Ahmed}, \bibinfo{person}{Yusra
  Saher}, {and} \bibinfo{person}{Ali~Raza Jafri}.}
  \bibinfo{year}{2014}\natexlab{}.
\newblock \showarticletitle{Assistive Glove for Pakistani Sign Language
  Translation}. In \bibinfo{booktitle}{{\em Multi-Topic Conference (INMIC),
  2014 IEEE 17th International}}. IEEE, \bibinfo{pages}{173--176}.
\newblock


\bibitem[\protect\citeauthoryear{Kim, Wagner, Rehm, and Andr{\'e}}{Kim
  et~al\mbox{.}}{2008}]%
        {kim2008bi}
\bibfield{author}{\bibinfo{person}{Jonghwa Kim}, \bibinfo{person}{Johannes
  Wagner}, \bibinfo{person}{Matthias Rehm}, {and} \bibinfo{person}{Elisabeth
  Andr{\'e}}.} \bibinfo{year}{2008}\natexlab{}.
\newblock \showarticletitle{Bi-channel sensor fusion for automatic sign
  language recognition}. In \bibinfo{booktitle}{{\em Automatic Face \& Gesture
  Recognition, 2008. FG'08. 8th IEEE International Conference on}}. IEEE,
  \bibinfo{pages}{1--6}.
\newblock


\bibitem[\protect\citeauthoryear{Kosmidou and Hadjileontiadis}{Kosmidou and
  Hadjileontiadis}{2009}]%
        {kosmidou2009sign}
\bibfield{author}{\bibinfo{person}{Vasiliki~E Kosmidou} {and}
  \bibinfo{person}{Leontios~J Hadjileontiadis}.}
  \bibinfo{year}{2009}\natexlab{}.
\newblock \showarticletitle{Sign language recognition using intrinsic-mode
  sample entropy on sEMG and accelerometer data}.
\newblock \bibinfo{journal}{{\em IEEE transactions on biomedical
  engineering\/}} \bibinfo{volume}{56}, \bibinfo{number}{12}
  (\bibinfo{year}{2009}), \bibinfo{pages}{2879--2890}.
\newblock


\bibitem[\protect\citeauthoryear{Li, Chen, Tian, Zhang, Wang, and Yang}{Li
  et~al\mbox{.}}{2010}]%
        {li2010automatic}
\bibfield{author}{\bibinfo{person}{Yun Li}, \bibinfo{person}{Xiang Chen},
  \bibinfo{person}{Jianxun Tian}, \bibinfo{person}{Xu Zhang},
  \bibinfo{person}{Kongqiao Wang}, {and} \bibinfo{person}{Jihai Yang}.}
  \bibinfo{year}{2010}\natexlab{}.
\newblock \showarticletitle{Automatic recognition of sign language subwords
  based on portable accelerometer and EMG sensors}. In \bibinfo{booktitle}{{\em
  International Conference on Multimodal Interfaces and the Workshop on Machine
  Learning for Multimodal Interaction}}. ACM, \bibinfo{pages}{17}.
\newblock


\bibitem[\protect\citeauthoryear{Li, Chen, Zhang, Wang, and Wang}{Li
  et~al\mbox{.}}{2012}]%
        {li2012sign}
\bibfield{author}{\bibinfo{person}{Yun Li}, \bibinfo{person}{Xiang Chen},
  \bibinfo{person}{Xu Zhang}, \bibinfo{person}{Kongqiao Wang}, {and}
  \bibinfo{person}{Z~Jane Wang}.} \bibinfo{year}{2012}\natexlab{}.
\newblock \showarticletitle{A sign-component-based framework for Chinese sign
  language recognition using accelerometer and sEMG data}.
\newblock \bibinfo{journal}{{\em IEEE transactions on biomedical
  engineering\/}} \bibinfo{volume}{59}, \bibinfo{number}{10}
  (\bibinfo{year}{2012}), \bibinfo{pages}{2695--2704}.
\newblock


\bibitem[\protect\citeauthoryear{Liddell}{Liddell}{2003}]%
        {liddell2003grammar}
\bibfield{author}{\bibinfo{person}{Scott~K Liddell}.}
  \bibinfo{year}{2003}\natexlab{}.
\newblock \bibinfo{booktitle}{{\em Grammar, gesture, and meaning in American
  Sign Language}}.
\newblock \bibinfo{publisher}{Cambridge University Press}.
\newblock


\bibitem[\protect\citeauthoryear{Marin, Dominio, and Zanuttigh}{Marin
  et~al\mbox{.}}{2014}]%
        {marin2014hand}
\bibfield{author}{\bibinfo{person}{Giulio Marin}, \bibinfo{person}{Fabio
  Dominio}, {and} \bibinfo{person}{Pietro Zanuttigh}.}
  \bibinfo{year}{2014}\natexlab{}.
\newblock \showarticletitle{Hand gesture recognition with leap motion and
  kinect devices}. In \bibinfo{booktitle}{{\em Image Processing (ICIP), 2014
  IEEE International Conference on}}. IEEE, \bibinfo{pages}{1565--1569}.
\newblock


\bibitem[\protect\citeauthoryear{McGraw, Prabhavalkar, Alvarez, Arenas, Rao,
  Rybach, Alsharif, Sak, Gruenstein, Beaufays, and Parada}{McGraw
  et~al\mbox{.}}{2016}]%
        {mcgraw2016personalized}
\bibfield{author}{\bibinfo{person}{Ian McGraw}, \bibinfo{person}{Rohit
  Prabhavalkar}, \bibinfo{person}{Raziel Alvarez},
  \bibinfo{person}{Montse~Gonzalez Arenas}, \bibinfo{person}{Kanishka Rao},
  \bibinfo{person}{David Rybach}, \bibinfo{person}{Ouais Alsharif},
  \bibinfo{person}{Ha{\c{s}}im Sak}, \bibinfo{person}{Alexander Gruenstein},
  \bibinfo{person}{Fran{\c{c}}oise Beaufays}, {and} \bibinfo{person}{Carolina
  Parada}.} \bibinfo{year}{2016}\natexlab{}.
\newblock \showarticletitle{Personalized speech recognition on mobile devices}.
  In \bibinfo{booktitle}{{\em Acoustics, Speech and Signal Processing (ICASSP),
  2016 IEEE International Conference on}}. IEEE, \bibinfo{pages}{5955--5959}.
\newblock


\bibitem[\protect\citeauthoryear{Melgarejo, Zhang, Ramanathan, and
  Chu}{Melgarejo et~al\mbox{.}}{2014}]%
        {melgarejo2014leveraging}
\bibfield{author}{\bibinfo{person}{Pedro Melgarejo}, \bibinfo{person}{Xinyu
  Zhang}, \bibinfo{person}{Parameswaran Ramanathan}, {and}
  \bibinfo{person}{David Chu}.} \bibinfo{year}{2014}\natexlab{}.
\newblock \showarticletitle{Leveraging directional antenna capabilities for
  fine-grained gesture recognition}. In \bibinfo{booktitle}{{\em Proceedings of
  the 2014 ACM International Joint Conference on Pervasive and Ubiquitous
  Computing}}. ACM, \bibinfo{pages}{541--551}.
\newblock


\bibitem[\protect\citeauthoryear{Mokaya, Lucas, Noh, and Zhang}{Mokaya
  et~al\mbox{.}}{2016}]%
        {mokaya2016burnout}
\bibfield{author}{\bibinfo{person}{Frank Mokaya}, \bibinfo{person}{Roland
  Lucas}, \bibinfo{person}{Hae~Young Noh}, {and} \bibinfo{person}{Pei Zhang}.}
  \bibinfo{year}{2016}\natexlab{}.
\newblock \showarticletitle{Burnout: a wearable system for unobtrusive skeletal
  muscle fatigue estimation}. In \bibinfo{booktitle}{{\em Information
  Processing in Sensor Networks (IPSN), 2016 15th ACM/IEEE International
  Conference on}}. IEEE, \bibinfo{pages}{1--12}.
\newblock


\bibitem[\protect\citeauthoryear{Nguyen, Alqurashi, Raghebi, Banaei-kashani,
  Halbower, and Vu}{Nguyen et~al\mbox{.}}{2016}]%
        {nguyen2016lightweight}
\bibfield{author}{\bibinfo{person}{Anh Nguyen}, \bibinfo{person}{Raghda
  Alqurashi}, \bibinfo{person}{Zohreh Raghebi}, \bibinfo{person}{Farnoush
  Banaei-kashani}, \bibinfo{person}{Ann~C Halbower}, {and} \bibinfo{person}{Tam
  Vu}.} \bibinfo{year}{2016}\natexlab{}.
\newblock \showarticletitle{A Lightweight And Inexpensive In-ear Sensing System
  For Automatic Whole-night Sleep Stage Monitoring}. In
  \bibinfo{booktitle}{{\em Proceedings of the 14th ACM Conference on Embedded
  Network Sensor Systems CD-ROM}}. ACM, \bibinfo{pages}{230--244}.
\newblock


\bibitem[\protect\citeauthoryear{Nirjon, Dickerson, Li, Asare, Stankovic, Hong,
  Zhang, Jiang, Shen, and Zhao}{Nirjon et~al\mbox{.}}{2012}]%
        {nirjon2012musicalheart}
\bibfield{author}{\bibinfo{person}{Shahriar Nirjon}, \bibinfo{person}{Robert~F
  Dickerson}, \bibinfo{person}{Qiang Li}, \bibinfo{person}{Philip Asare},
  \bibinfo{person}{John~A Stankovic}, \bibinfo{person}{Dezhi Hong},
  \bibinfo{person}{Ben Zhang}, \bibinfo{person}{Xiaofan Jiang},
  \bibinfo{person}{Guobin Shen}, {and} \bibinfo{person}{Feng Zhao}.}
  \bibinfo{year}{2012}\natexlab{}.
\newblock \showarticletitle{Musicalheart: A hearty way of listening to music}.
  In \bibinfo{booktitle}{{\em Proceedings of the 10th ACM Conference on
  Embedded Network Sensor Systems}}. ACM, \bibinfo{pages}{43--56}.
\newblock


\bibitem[\protect\citeauthoryear{Praveen, Karanth, and Megha}{Praveen
  et~al\mbox{.}}{2014}]%
        {praveen2014sign}
\bibfield{author}{\bibinfo{person}{Nikhita Praveen}, \bibinfo{person}{Naveen
  Karanth}, {and} \bibinfo{person}{MS Megha}.} \bibinfo{year}{2014}\natexlab{}.
\newblock \showarticletitle{Sign language interpreter using a smart glove}. In
  \bibinfo{booktitle}{{\em Advances in Electronics, Computers and
  Communications (ICAECC), 2014 International Conference on}}. IEEE,
  \bibinfo{pages}{1--5}.
\newblock


\bibitem[\protect\citeauthoryear{Savitzky and Golay}{Savitzky and
  Golay}{1964}]%
        {savitzky1964smoothing}
\bibfield{author}{\bibinfo{person}{Abraham Savitzky} {and}
  \bibinfo{person}{Marcel~JE Golay}.} \bibinfo{year}{1964}\natexlab{}.
\newblock \showarticletitle{Smoothing and differentiation of data by simplified
  least squares procedures.}
\newblock \bibinfo{journal}{{\em Analytical chemistry\/}} \bibinfo{volume}{36},
  \bibinfo{number}{8} (\bibinfo{year}{1964}), \bibinfo{pages}{1627--1639}.
\newblock


\bibitem[\protect\citeauthoryear{Schuster and Paliwal}{Schuster and
  Paliwal}{1997}]%
        {schuster1997bidirectional}
\bibfield{author}{\bibinfo{person}{Mike Schuster} {and}
  \bibinfo{person}{Kuldip~K Paliwal}.} \bibinfo{year}{1997}\natexlab{}.
\newblock \showarticletitle{Bidirectional recurrent neural networks}.
\newblock \bibinfo{journal}{{\em IEEE Transactions on Signal Processing\/}}
  \bibinfo{volume}{45}, \bibinfo{number}{11} (\bibinfo{year}{1997}),
  \bibinfo{pages}{2673--2681}.
\newblock


\bibitem[\protect\citeauthoryear{Socher, Lin, Manning, and Ng}{Socher
  et~al\mbox{.}}{2011}]%
        {socher2011parsing}
\bibfield{author}{\bibinfo{person}{Richard Socher}, \bibinfo{person}{Cliff~C
  Lin}, \bibinfo{person}{Chris Manning}, {and} \bibinfo{person}{Andrew~Y Ng}.}
  \bibinfo{year}{2011}\natexlab{}.
\newblock \showarticletitle{Parsing natural scenes and natural language with
  recursive neural networks}. In \bibinfo{booktitle}{{\em Proceedings of the
  28th international conference on machine learning (ICML-11)}}.
  \bibinfo{pages}{129--136}.
\newblock


\bibitem[\protect\citeauthoryear{Starner, Weaver, and Pentland}{Starner
  et~al\mbox{.}}{1998}]%
        {starner1998real}
\bibfield{author}{\bibinfo{person}{Thad Starner}, \bibinfo{person}{Joshua
  Weaver}, {and} \bibinfo{person}{Alex Pentland}.}
  \bibinfo{year}{1998}\natexlab{}.
\newblock \showarticletitle{Real-time american sign language recognition using
  desk and wearable computer based video}.
\newblock \bibinfo{journal}{{\em IEEE Transactions on Pattern Analysis and
  Machine Intelligence\/}} \bibinfo{volume}{20}, \bibinfo{number}{12}
  (\bibinfo{year}{1998}), \bibinfo{pages}{1371--1375}.
\newblock


\bibitem[\protect\citeauthoryear{Sutskever, Vinyals, and Le}{Sutskever
  et~al\mbox{.}}{2014}]%
        {sutskever2014sequence}
\bibfield{author}{\bibinfo{person}{Ilya Sutskever}, \bibinfo{person}{Oriol
  Vinyals}, {and} \bibinfo{person}{Quoc~V Le}.}
  \bibinfo{year}{2014}\natexlab{}.
\newblock \showarticletitle{Sequence to sequence learning with neural
  networks}. In \bibinfo{booktitle}{{\em Advances in neural information
  processing systems}}. \bibinfo{pages}{3104--3112}.
\newblock


\bibitem[\protect\citeauthoryear{Tubaiz, Shanableh, and Assaleh}{Tubaiz
  et~al\mbox{.}}{2015}]%
        {tubaiz2015glove}
\bibfield{author}{\bibinfo{person}{Noor Tubaiz}, \bibinfo{person}{Tamer
  Shanableh}, {and} \bibinfo{person}{Khaled Assaleh}.}
  \bibinfo{year}{2015}\natexlab{}.
\newblock \showarticletitle{Glove-based continuous Arabic sign language
  recognition in user-dependent mode}.
\newblock \bibinfo{journal}{{\em IEEE Transactions on Human-Machine Systems\/}}
  \bibinfo{volume}{45}, \bibinfo{number}{4} (\bibinfo{year}{2015}),
  \bibinfo{pages}{526--533}.
\newblock


\bibitem[\protect\citeauthoryear{Uebersax, Gall, Van~den Bergh, and
  Van~Gool}{Uebersax et~al\mbox{.}}{2011}]%
        {uebersax2011real}
\bibfield{author}{\bibinfo{person}{Dominique Uebersax},
  \bibinfo{person}{Juergen Gall}, \bibinfo{person}{Michael Van~den Bergh},
  {and} \bibinfo{person}{Luc Van~Gool}.} \bibinfo{year}{2011}\natexlab{}.
\newblock \showarticletitle{Real-time sign language letter and word recognition
  from depth data}. In \bibinfo{booktitle}{{\em Computer Vision Workshops (ICCV
  Workshops), 2011 IEEE International Conference on}}. IEEE,
  \bibinfo{pages}{383--390}.
\newblock


\bibitem[\protect\citeauthoryear{Von~Agris, Zieren, Canzler, Bauer, and
  Kraiss}{Von~Agris et~al\mbox{.}}{2008}]%
        {von2008recent}
\bibfield{author}{\bibinfo{person}{Ulrich Von~Agris}, \bibinfo{person}{J{\"o}rg
  Zieren}, \bibinfo{person}{Ulrich Canzler}, \bibinfo{person}{Britta Bauer},
  {and} \bibinfo{person}{Karl-Friedrich Kraiss}.}
  \bibinfo{year}{2008}\natexlab{}.
\newblock \showarticletitle{Recent developments in visual sign language
  recognition}.
\newblock \bibinfo{journal}{{\em Universal Access in the Information
  Society\/}} \bibinfo{volume}{6}, \bibinfo{number}{4} (\bibinfo{year}{2008}),
  \bibinfo{pages}{323--362}.
\newblock


\bibitem[\protect\citeauthoryear{Wang, Wu, and Ni}{Wang et~al\mbox{.}}{2017}]%
        {wang2017wifall}
\bibfield{author}{\bibinfo{person}{Yuxi Wang}, \bibinfo{person}{Kaishun Wu},
  {and} \bibinfo{person}{Lionel~M Ni}.} \bibinfo{year}{2017}\natexlab{}.
\newblock \showarticletitle{Wifall: Device-free fall detection by wireless
  networks}.
\newblock \bibinfo{journal}{{\em IEEE Transactions on Mobile Computing\/}}
  \bibinfo{volume}{16}, \bibinfo{number}{2} (\bibinfo{year}{2017}),
  \bibinfo{pages}{581--594}.
\newblock


\bibitem[\protect\citeauthoryear{Wu, Tian, Sun, Estevez, and Jafari}{Wu
  et~al\mbox{.}}{2015}]%
        {wu2015real}
\bibfield{author}{\bibinfo{person}{Jian Wu}, \bibinfo{person}{Zhongjun Tian},
  \bibinfo{person}{Lu Sun}, \bibinfo{person}{Leonardo Estevez}, {and}
  \bibinfo{person}{Roozbeh Jafari}.} \bibinfo{year}{2015}\natexlab{}.
\newblock \showarticletitle{Real-time American sign language recognition using
  wrist-worn motion and surface EMG sensors}. In \bibinfo{booktitle}{{\em
  Wearable and Implantable Body Sensor Networks (BSN), 2015 IEEE 12th
  International Conference on}}. IEEE, \bibinfo{pages}{1--6}.
\newblock


\bibitem[\protect\citeauthoryear{Zafrulla, Brashear, Hamilton, and
  Starner}{Zafrulla et~al\mbox{.}}{2010}]%
        {zafrulla2010novel}
\bibfield{author}{\bibinfo{person}{Zahoor Zafrulla}, \bibinfo{person}{Helene
  Brashear}, \bibinfo{person}{Harley Hamilton}, {and} \bibinfo{person}{Thad
  Starner}.} \bibinfo{year}{2010}\natexlab{}.
\newblock \showarticletitle{A novel approach to american sign language (asl)
  phrase verification using reversed signing}. In \bibinfo{booktitle}{{\em
  Computer Vision and Pattern Recognition Workshops (CVPRW), 2010 IEEE Computer
  Society Conference on}}. IEEE, \bibinfo{pages}{48--55}.
\newblock


\bibitem[\protect\citeauthoryear{Zhu, Lan, Xing, Zeng, Li, Shen, and Xie}{Zhu
  et~al\mbox{.}}{2016}]%
        {Zhu2016}
\bibfield{author}{\bibinfo{person}{Wentao Zhu}, \bibinfo{person}{Cuiling Lan},
  \bibinfo{person}{Junliang Xing}, \bibinfo{person}{Wenjun Zeng},
  \bibinfo{person}{Yanghao Li}, \bibinfo{person}{Li Shen}, {and}
  \bibinfo{person}{Xiaohui Xie}.} \bibinfo{year}{2016}\natexlab{}.
\newblock \showarticletitle{Co-occurrence Feature Learning for Skeleton Based
  Action Recognition Using Regularized Deep LSTM Networks}. In
  \bibinfo{booktitle}{{\em Proceedings of the Thirtieth AAAI Conference on
  Artificial Intelligence}} {\em (\bibinfo{series}{AAAI'16})}.
  \bibinfo{publisher}{AAAI Press}, \bibinfo{pages}{3697--3703}.
\newblock
\showURL{%
\url{http://dl.acm.org/citation.cfm?id=3016387.3016423}}


\end{thebibliography}

\end{document}